\pdfoutput=1

\documentclass[11pt]{article}

\usepackage{acl}
\usepackage{times}
\usepackage{latexsym}
\usepackage[T1]{fontenc}
\usepackage{soul}
\usepackage{multirow}
\usepackage{makecell}
\usepackage{boxedminipage}
\usepackage[utf8]{inputenc}
\usepackage{microtype}
\usepackage{graphicx}
\usepackage{soul}
\usepackage{float}
\usepackage{comment}
\usepackage{adjustbox}
\usepackage{booktabs}
\usepackage{amsmath}
\usepackage{mathabx}
\usepackage[shortlabels]{enumitem}
\usepackage[normalem]{ulem}
\usepackage{xspace}
\usepackage{array}
\usepackage{framed}
\usepackage{subcaption}
\usepackage{xcolor}
\usepackage{multicol}

\newcommand{\task}[1]{$\mathcal{T}_{\textnormal{#1}}$}

\newcolumntype{L}[1]{>{\raggedright\let\newline\\\arraybackslash\hspace{0pt}}m{#1}}
\newcolumntype{C}[1]{>{\centering\let\newline\\\arraybackslash\hspace{0pt}}m{#1}}

\definecolor{purple}{rgb}{0.5,0,1}
\definecolor{dcyan}{rgb}{0.2,0.6,0.5}
\definecolor{darkgreen}{rgb}{0,200,0}
\definecolor{darkorange}{rgb}{138, 50, 0}
\definecolor{light-gray}{gray}{0.95} 
\definecolor{darkgreen}{RGB}{0,140,0}
\definecolor{darkred}{RGB}{200,0,0}
\definecolor{lightgreen}{RGB}{231,255,219}
\definecolor{lightred}{RGB}{252,231,234}
\definecolor{lightyellow}{RGB}{250,253,191}
\definecolor{DarkRed}{RGB}{130,25,0}

\newcommand{\changed}[1]{{#1}}

\newcommand{\changednew}[1]{{#1}}

\newcommand{\cha}[1]{{\color{orange}#1}}

\providecommand{\hanna}[1]{
    {\protect\color{blue}{[Hanna: #1]}}
}

\newcommand{\name}{\textsc{Natural Instructions}}

\newcommand{\swaroop}[1]{}

\newcommand{\setOf}[1]{\left\lbrace #1 \right\rbrace}
\newcommand{\I}[2]{I^{ \textnormal{\tiny #2}}_{#1}}

\title{Learn from Instructions instead of Learning just from Data Samples}
\title{\name{}: A Study of Language-Models in According to Instructions}
\title{Can Language Models Follow Instructions?}
\title{\name{}: Steps Towards Models that \\ Follow Instructions}
\title{\name{}: Towards Models that Follow Instructions}
\title{\name{}: Inter-Task Generalization \\ 
via Natural Language Instructions }
\title{\name{}: Generalization to New Tasks from Instructions }
\title{\name{}: Learning New Tasks from Instructions }
\title{\name{}: Can We Learn New Tasks from Instructions?}
\title{\name{}: Assessing Generalization to New Tasks from Instructions }
\title{\name{}: Benchmarking Generalization \\ 
to New Tasks from Natural Language Instructions }
\title{\name{}: Benchmarking Cross-Task Generalization \\ 
via Natural Language Instructions }
\title{Cross-Task Generalization \\ 
via Natural Language Instructions }
\title{Improving Generalization to Unseen Tasks \\ 
via Natural Language Instructions }
\title{Generalization to Unseen Tasks via Natural Language Instructions }
\title{\changed{Task-Level} Generalization \\ via Natural Language Crowdsourcing Instructions}
\title{
\vspace*{-0.5in}
{{\small \hfill ACL 2022}\\
\vspace*{.25in}} 
Cross-Task Generalization \\ via Natural Language Crowdsourcing Instructions}

\author{
Swaroop Mishra$^{3}$\thanks{~~Work done while interning at Allen Institute for AI.} $\;$ Daniel Khashabi$^{1}$ $\;$ \textbf{Chitta Baral}$^{3}$ $\;$ \textbf{Hannaneh Hajishirzi}$^{1,2}$ $\;$ 
\\\\
 $^1$Allen Institute for AI \; $^2$University of Washington \; 
 $^3$Arizona State University 
}

\newcommand{\egbox}[1]{
\smallskip
\noindent
\fbox{
        \parbox{0.95\linewidth}{
        \vspace{0.5ex}#1\vspace{0.5ex}}
      }
\smallskip
}

\begin{document}
\maketitle

\begin{abstract} 
 Humans (e.g., crowdworkers) have a remarkable ability in solving different tasks, by simply reading textual \emph{instructions} that define them and looking at a few examples. 
\changednew{Despite the success of the conventional supervised learning on individual datasets, such models}
often struggle with generalization across tasks (e.g., a question-answering system cannot solve  classification tasks).
 A long-standing challenge in AI is to build a model that learns a new task by understanding the human-readable \emph{instructions} that define it.
 To study this, we introduce \name,  a dataset of  
 61  distinct tasks, their human-authored instructions, and 193$k$ task instances \changednew{(input-output pairs)}. The instructions are obtained from crowdsourcing instructions used to create  existing NLP datasets and  mapped to a unified schema.  
 Using this meta-dataset, we measure cross-task generalization by training models on {\emph{\color{blue} seen}}  tasks and measuring generalization to the remaining {\emph{\color{red} unseen}} ones.
 We adopt generative pre-trained language models to encode task-specific instructions along with input and generate task output. 
 Our results indicate that models \emph{benefit from instructions} when evaluated in terms of generalization to unseen tasks \changednew{(19\% better for models utilizing instructions)}. 
 These models, however, are far behind 
 \changednew{an estimated performance upperbound,}
 indicating significant room for more progress in this direction.\footnote{
Dataset is available at \cha{\url{https://instructions.apps.allenai.org}}
}
\end{abstract}

\section{Introduction}

\begin{figure}[t]
    \centering
    \includegraphics[scale=0.9, trim=0.75cm 0.8cm 0cm 1.0cm,clip=false]{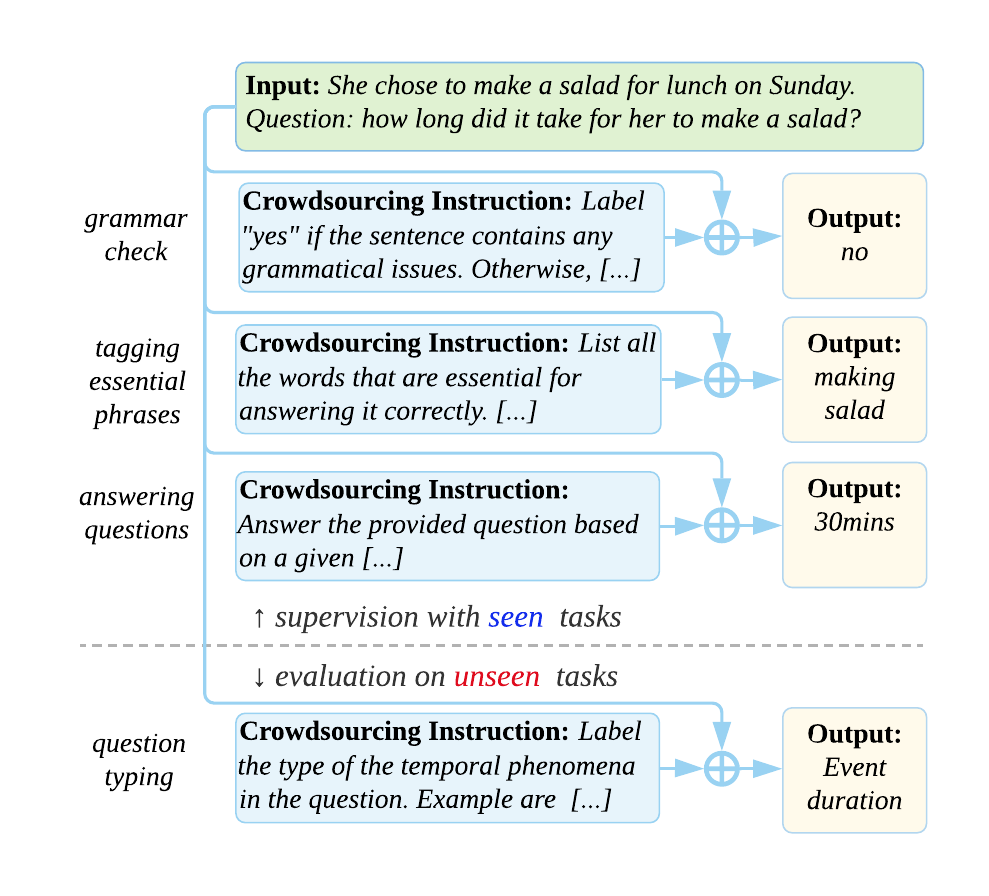}
    \caption{
        We construct the \name{} dataset from crowdsourcing instructions and instances of different NLP datasets.  We study if models can learn from  {\emph{\color{blue} seen}}  tasks and generalize to  {\emph{\color{red} unseen}} tasks given their natural crowdsourcing instructions. 
    }
    \label{fig:teaster}
\end{figure}

We have witnessed great progress in solving many NLP datasets through fine-tuning pre-trained language models (LMs)~\cite{peters2018deep,brown2020language}. 
More recent studies show tremendous promise in  generalization \emph{within} the set of observed tasks through multi-task training and unified encoding~\cite{khashabi2020unifiedqa,aghajanyan2021muppet}.  
However, cross-task generalization -- \emph{generalization} to  \emph{unseen} tasks -- has generally remained under-explored.
For example, can we supervise a model with instances of grammar checking or question answering tasks, yet expect it to solve a different task like question typing (Fig.\ref{fig:teaster}).
Evidently, humans are capable of such generalizations; an average human can follow natural language \emph{instructions} to solve a variety of problems, as evident by the success of crowdsourcing platforms (also argued in~\citet{efrat2020turking}). In this paper, we study if models can generalize to  \emph{unseen} tasks given their 
crowdsourcing instructions (Fig.\ref{fig:teaster}).

\begin{figure*}
     \centering
     \begin{subfigure}[b]{0.49\textwidth}
         \centering
            \small
        \resizebox{\linewidth}{!}{
        \begin{tabular}{ccc}
            \toprule 
             Task   & \makecell{Instance-Level\\Generalization}  & \makecell{Task-Level\\Generalization} \\
             \midrule 
             \makecell{Training\\data}  & $X^{\color{darkgreen} \text{train}}, Y^{\color{darkgreen} \text{train}}$ & \makecell{$(I_t, X_t^{{\color{darkgreen} \text{train}}}, Y_t^{{\color{darkgreen} \text{train}}})$ \\ $t \in \text{\task{\color{blue} seen}}  $ \\ }  \\ 
             \midrule 
             Evaluation &  \makecell{ $x \rightarrow y$ \vspace{0.2cm} \\ where: \\  $(x, y) \in (X^{ \color{purple} \text{test}}, Y^{ \color{purple} \text{test}})$  \vspace{0.3cm} }  & \makecell{$(x, I_t) \rightarrow y$ \vspace{0.2cm} \\  where: \\ $(x, y) \in (X_t^{ {\color{purple} \text{test}}}, Y_t^{{\color{purple} \text{test}}})$ \\ $t \in$ \task{\color{red} unseen} } \\ 
            \bottomrule
        \end{tabular}
        }
        \caption{
            A comparison of \emph{task} vs \emph{instance}-level generalization 
            $I_t$, $X_t$ and $Y_t$ indicate natural language instructions, input, and output sets respectively for task $t$.
            In the conventional setup, training and evaluation are done on the instances of the same task. 
            However, in task-level generalization, a model is expected to generalize to  {\color{red} unseen} tasks, where \task{\color{red} unseen} $\cap$ \task{\color{blue} seen}$ = \emptyset $. 
        }
        \label{tab:comparison}
     \end{subfigure}
     \hfill
     \begin{subfigure}[b]{0.49\textwidth}
         \centering
         \includegraphics[scale=0.64,trim=0.2cm 0cm 0cm 1cm,clip=false]{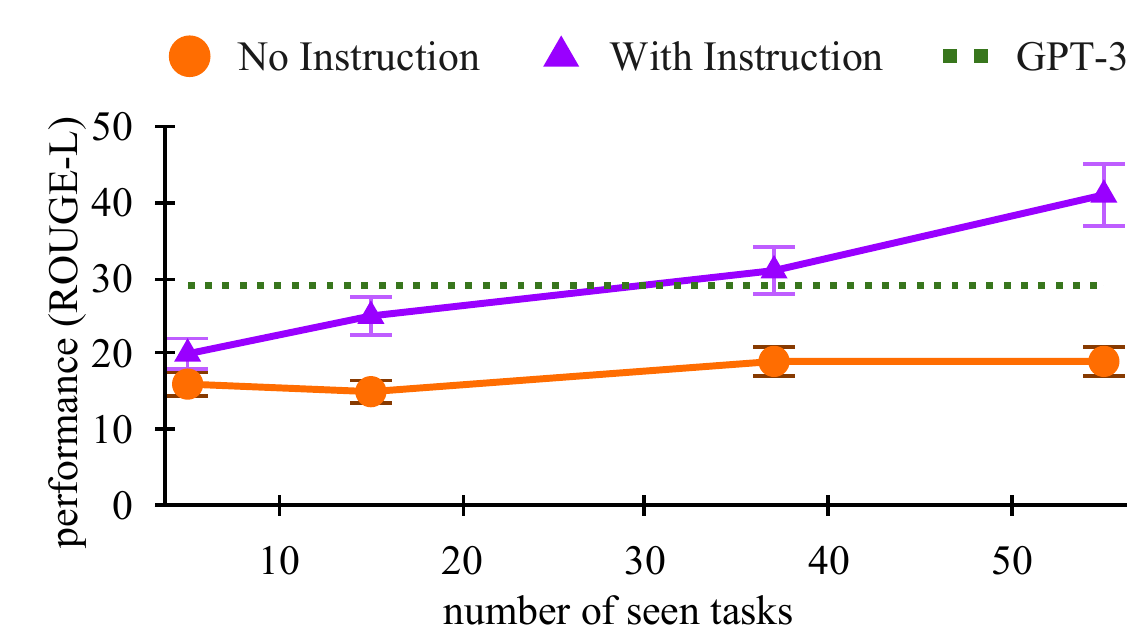}
         \caption{BART evaluation on {\emph{unseen}} tasks ($y$-axis is perf. on \task{unseen}) when supervised with {\emph{seen}} tasks ($x$-axis is $|$\task{seen}$|$). 
            \changed{
             A model using {\color{purple}instructions} ($I_t$) consistently improves with more observed tasks. In contrast, models with {\color{orange} no access to the instructions} show no sign of improved generalization. 
             }
            \changed{Details in \S\ref{subsec:supervision:size:experiment}.}
        }
        \label{fig:scaling:tasks}
    \end{subfigure}
    \caption{The formal definition of generalization to unseen tasks (a) and a summary of its empirical outcome (b).  }
\end{figure*}

We build \name, a dataset consisting of {\it natural} crowdsourcing instructions for various tasks and their instances. 
Training on {\it seen} tasks $\text{\task{\color{blue} seen}}$ in our dataset, we  build a model that learns to follow natural instructions that define a task and perform tasks  (i.e., mapping input to output).
Testing on \emph{unseen} tasks \text{\task{\color{red} unseen}}, we evaluate if the model can perform {\it unseen} tasks  solely from their instructions and without any task-specific labeled data   (Table~\ref{tab:comparison}; right).  
In contrast to the instance-level generalization (Table~\ref{tab:comparison}; left), our model uses instruction as additional input, and evaluations are done on tasks that were not observed in the training stage. 

\changed{
We compile \name{} from task instructions written by researchers for crowdsourcing existing NLP datasets. 
Such crowdsourcing instructions often elaborate a variety of details about how a task should (and should not) be done.  
To provide a systematic study of various elements of crowdsourcing instructions, we map them
}
to a unified {\it schema} to cover the most important elements of task descriptions --- such as definition, constraints, positive and negative examples. 
We collect tasks in  \name{}  as minimal stand-alone steps provided to crowdworkers to complete a downstream NLP task. 
For example, tasks collected from  
\changed{QASC~\cite{khot2020qasc} include sub-tasks about generating topic words or combining facts, as well as answering multi-hop questions. 
Therefore our dataset not only contains typical downstream tasks in NLP, but also the intermediate subtasks that are not well-represented in the common benchmarks. 
}
The unified schema and the collection of minimal subtasks enable training LMs that can generalize across different tasks by learning from instructions.
In total, our dataset consists of 61 distinct NLP tasks and $193k$ instances.

Our experimental results indicate that LMs learn to leverage natural language instructions as they show improved generalization to new
tasks. 
For example, a BART~\cite{lewis2019bart}  achieves a 19\% gain in terms of cross-task generalization compared to a model not using instructions
(\S\ref{sec:experiments}). 
Importantly, LMs can generalize better to unseen tasks if they observe more tasks in training (Fig.\ref{fig:scaling:tasks}). 
This upward trajectory suggests the potential for stronger cross-task generalizable models upon scaling up the diversity of tasks represented in a meta-dataset of task instructions. 
Despite the benefits of instructions, we observe a sizable gap between models' generalization and their estimated upperbounds (\ref{subsec:task-specific}), encouraging the community to work on this challenging problem.

\vspace{.1cm}
\noindent\textbf{Contributions:} In summary, the contributions of this work are as follows: 
(a) we introduce \name{}, a  dataset of human-authored instructions  curated from existing well-known datasets mapped to a unified schema, providing training and evaluation data for learning from instructions;
(b) we build models that can encode instructions and show: 
(b.1) the benefit of cross-task generalization by leveraging instructions; 
(b.2) the importance of different elements of instructions in the performance;  
(b.3) noteworthy headroom for improvement on our benchmark, which hopefully will motivate further work in this direction. 

\section{Related Works}
\label{sec:related:work}

\changed{
\vspace{-.2cm} \paragraph{Learning from instructions.}
There is recent literature on the extent to which models follow language instructions~~\cite{hase2021can,ye2021zero,Gupta2021TowardsGP,Zhong2021AdaptingLM}.
For example, \citet{efrat2020turking} examine if language models can follow crowdsourcing  instructions with no further training. On the contrary, our work is pursuing a fundamentally different goal: creating a dataset of crowdsourcing instructions and task instances and formulating cross-task generalization by training models on seen tasks and measuring generalization to the remaining unseen ones.
\citet{weller-etal-2020-learning} construct a crowdsourced dataset with short question-like task descriptions.  
Compared to this work, our instructions are longer, more complex and natural since they were used to collect datasets through crowdsourcing. 

PromptSource and FLAN~\cite{wei2021finetuned,sanh2021multitask} are two concurrent works that pursue a similar goal as ours. 
A key difference between our work to these works is in terms of data collection strategy. 
Our work uses natural instructions created by NLP researchers before the dataset instances were created by crowd workers, and hence it contains the complete definition of each task (definition, things to avoid, negative examples, etc.). 
On the other hand, instructions in the concurrent work are collected retroactively based on the already-available task instances. 
Our  {\it natural} instructions enable evaluating models on how they learn tasks given different elements of task descriptions. (See \S\ref{subsec:promptsource} for further comparisons.) 
Nevertheless, we believe that all these approaches to constructing instructions and task categories are complementary and the community will benefit from considering both towards solving the challenging problem of cross-task generalization.

\vspace{-.2cm}\paragraph{Prompt engineering.}
Constructing effective discrete prompts for language models to perform NLP tasks is an active area of research~\cite{schick2020few,reynolds2021prompt,liu2021pre}. 
Such prompts are often extremely short and may not include a complete definition of complex tasks. 
In contrast, our instructions encode detailed instructions as they were used to collect the datasets. 
Moreover, the goals are different:
Most prompt-engineering approaches seek prompts with higher performance on a particular task, 
typically through assumptions about their target task which make them non-trivial to generalize to any other task. 
However, our introduced meta dataset enables the measurement of generalization to unseen tasks.

\vspace{-.2cm}\paragraph{Beyond standard multi-task learning.}
Multi-task learning is a long-standing goal for AI~\cite{caruana1997multitask} and has led to successful models that can support a wider range of tasks
~\cite{mccann2018natural,raffel2020exploring,khashabi2020unifiedqa,mishra2020towards,aghajanyan2021muppet,ye2021crossfit}.
Most of the conventional setups in the multi-tasking literature  evaluate on instances that belong to the tasks that are seen, i.e., their labeled instances were observed during training (1st column of Table~\ref{tab:comparison}). 
We augment this setup by 
introducing natural language instructions which enable our models to bridge to tasks that were not seen during training. 
}

\changed{
\section{Defining Cross-Task Generalization}
\label{subsec:input:output}
Here we formally define the problem setup for generalization across tasks. 
Each task $t$ consists of input/output instances $(X_t, Y_t)$ and is described in terms of its natural language instructions $I_t$. 


%
%
\vspace{-.2cm}
\paragraph{Task-specific models.}
Standard supervised learning algorithms use task-specific labeled instances to learn a mapping from input $x$ to output $y$:  $M(x)=y$ for $(x,y)\in (X_t^{\text{train}}, Y_t^{\text{train}})$ and is evaluated on the test instances of the same (or similar) task $(X_t^{\text{test}}, Y_t^{\text{test}})$. We refer to this as the \emph{instance-level} generalization (Table~\ref{tab:comparison}; left).

\vspace{-.2cm}
\paragraph{Cross-task models.} 
In this setup, the goal is to learn a model $M$ that at inference obtains the output $y$ given the input $x$ and the task instruction $I_t$: $M(I_t, x) = y, \;  \mbox{for} \ (x,y)\in (X_t, Y_t)$.
In contrast to the task-specific models, no task-specific training data is used to learn the mapping $M$. We collect \name\ (\S\ref{sec:construction:natural:instructions}) to study this question: can a model be trained to follow instructions via training tasks \task{seen} and be generalized to follow instructions for a task $t' \in$ \task{unseen}. 
We refer to this as a \emph{task}-level generalization (Table~\ref{tab:comparison}; right). 
}

\section{\name{}}
\label{sec:construction:natural:instructions}

\begin{figure}[t]
    \centering
    \includegraphics[scale=0.70, trim=0.45cm 0.8cm 0cm 0.99cm]{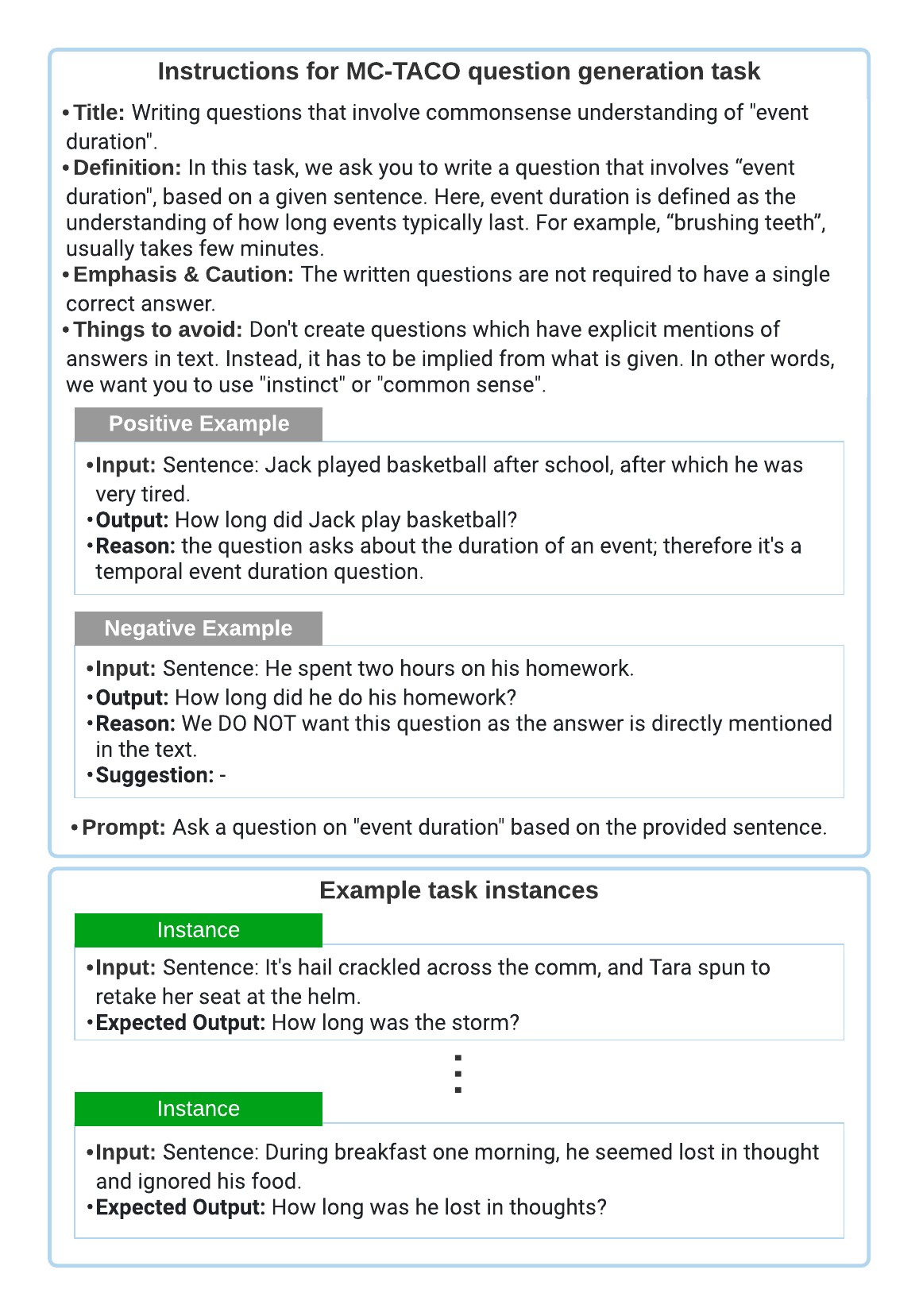}
    \caption{
        An example from our dataset.  
        Note that it follows the schema provided in Fig.\ref{fig:schema_plate}. See Fig~.\ref{fig:examplesfull} for more examples.
    }
    \label{fig:examples}
\end{figure}

\name{} consists of instructions that describe a task (e.g., question answering) and instances of that task (e.g., answers extracted for a given question). 
Fig.\ref{fig:examples} shows  an example instruction for the task of `generating questions that require an understanding of event duration' accompanied with positive and negative examples 
that contextualize the task. 
Here we introduce a schema for representing instructions  (\S\ref{subsec:schema}) and then describe how existing datasets (their crowdsourcing templates) are mapped into  our schema  (\S\ref{sec:mapping}). 



\subsection{Instruction Schema}
\label{subsec:schema}


Instructions used in crowdsourcing various datasets, are written by distinct authors for different purposes, and  they are different in a variety of ways (see  Appendix~\ref{appendix:analysis:templates} for their differences.) We introduce a unified schema (Fig.\ref{fig:schema_plate}) to consistently represent these diverse forms of instructions. 
Our instruction schema is the result of our pilot study conducted on a subset of  datasets. Below we describe the ingredients of this schema: 

\begin{figure}[t]
\centering
    \includegraphics[width=0.97\columnwidth,trim=0.35cm 0.8cm 0.5cm 1cm]{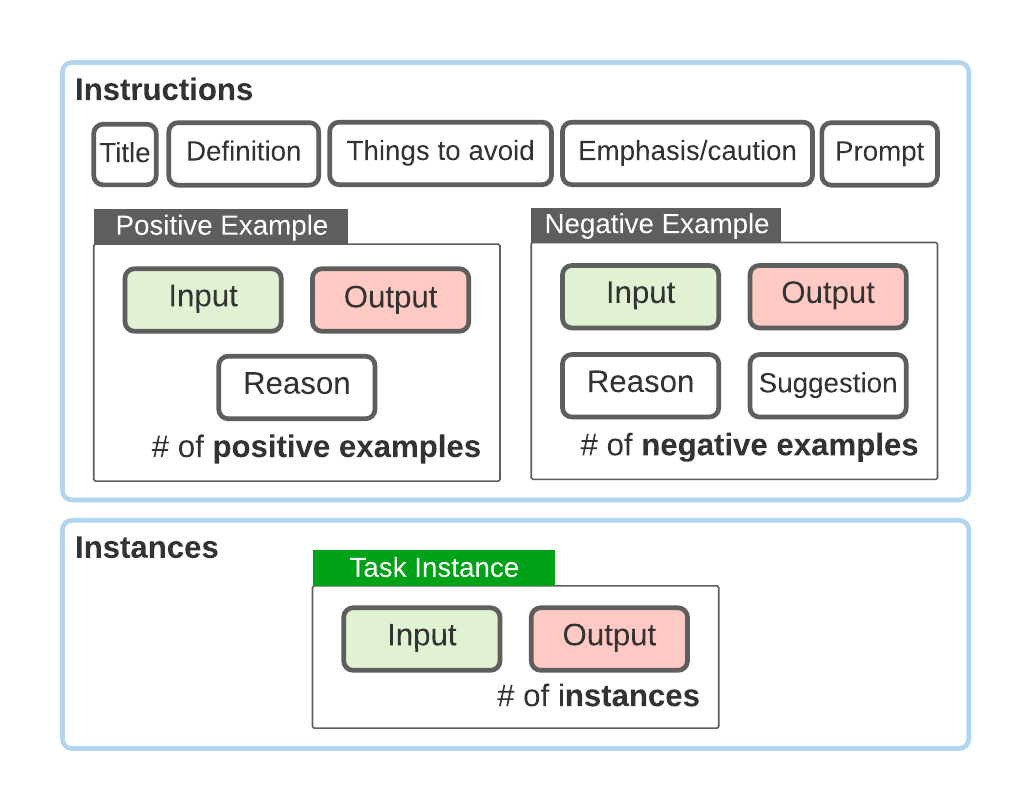}
  \caption{The schema used for representing instruction in \name{} (\S\ref{subsec:schema}), shown in plate notation.
  }
  \label{fig:schema_plate}
\end{figure}

\begin{itemize}[noitemsep,topsep=0pt,parsep=3pt,leftmargin=0.3cm]
    \item \underline{\textsc{Title}} provides a high-level description of a task and its associated skill (such as question generation, answer generation).
    \item  \underline{\textsc{Prompt}} is a single sentence command that often appears before the input instance and connects it to the instructions.
    \item \underline{\textsc{Definition}} provides the core detailed instructions for a task. 
    \item \underline{\textsc{Things to Avoid}} contain instructions regarding undesirable annotations that must be avoided. These help to define the scope of a task and the space of acceptable responses. 
    \item \underline{\textsc{Emphasis and Caution}} are short, but important statements highlighted in the crowdsourcing templates which were intended to be  emphasized or warned against. 
    \item \underline{\textsc{Positive Examples}} contain inputs/outputs similar to the input given to a worker/system and its expected output,  helping crowdworkers better understand a task~\cite{ali1981use}. 
    \item \underline{\textsc{Negative Examples}} contain inputs/outputs to emphasize \textsc{Things to Avoid} by providing examples that must not be produced. 
    \item \underline{\textsc{Reason}}  provides explanations behind why an example is positive or negative.
    \item \underline{\textsc{Suggestion}} contains suggestions on how a negative example could be modified to turn it into a positive example. 
\end{itemize}

    The next section describes the process of mapping the raw instructions (designed for crowdworkers) to our instruction schema.

\subsection{Constructing \name} 
\label{sec:mapping}



\subsubsection{Collecting Data}
\label{sec:datacollection}
\paragraph{Collecting raw instructions and instances.} 
    We use existing, widely adopted NLP benchmarks that are collected via crowdsourcing platforms and hence, come with crowdsourcing templates. 
    In the first step, we identified several datasets and engaged with their authors to get their crowdsourcing templates and raw data. 
This yields the following datasets: 
CosmosQA~\cite{huang2019cosmos}, 
DROP~\cite{dua2019drop}, 
Essential-Terms~\cite{khashabi2017learning}, 
MCTACO~\cite{zhou2019going}, 
MultiRC~\cite{khashabi2018looking}, 
QASC~\cite{khot2020qasc}, 
Quoref~\cite{dasigi2019quoref}, ROPES~\cite{lin2019reasoning} and
Winogrande~\cite{sakaguchi2020winogrande}.\footnote{
        We only focus on textual instructions and  avoid datasets that involve visual or auditory steps, mostly focusing on QA datasets that were available to the authors. 
} 
    
\vspace{-.2cm}
\paragraph{Splitting crowdsourcing instructions into minimal tasks.} 
Almost all the crowdworking instructions include sequences of steps to guide crowdworkers  in creating task instances.
For example, QASC and MCTACO include 7 and 19 steps in the data creation process, respectively. 
We divide crowdsourcing instructions into their underlying steps and generate multiple subtasks that are minimal and standalone.\footnote{
    We eliminate tasks that involve model-in-the-loop. 
} Table~\ref{tab:sample:tasks} shows subtasks extracted for Quoref and QASC. For example, the main task in Quoref is to answer a question given a context paragraph, but the crowdsourcing template consists of two sub-tasks of {\it question generation}  and {\it answer generation} with their separate instructions.  This process  results in a more consistent definition of tasks, enabling a successful mapping of instructions into our schema,  in contrast to the work of \citet{efrat2020turking} that uses crowdsourcing instructions as-is.

\begin{table}
    \centering
    \footnotesize 
    \begin{tabular}{ll}
        \toprule
        source dataset & task  \\
        \midrule
         \multirow{2}{*}{\makecell{Quoref\\ \cite{dasigi2019quoref} }} & question generation \\ 
            & answer generation \\ 
        \midrule
         \multirow{6}{*}{\makecell{QASC\\ \cite{khot2020qasc}} }  & topic word generation \\
                & fact generation \\ 
                & combining facts \\ 
                & question generation \\ 
                & answer generation \\ 
                & incorrect answer generation \\ 
        \bottomrule
    \end{tabular}
    \caption{
        Examples of the datasets and the tasks formed from them. 
        The extracted tasks are independent annotation assignments in the crowdsourcing templates of the datasets. 
        The complete list is in Table~\ref{tab:structure} in  Appendix. 
    }
    \label{tab:sample:tasks}
\end{table}

\begin{table}
    \footnotesize 
    \begin{tabular}{lcc}
        \toprule
        category  & \# of tasks  & \# of instances  \\
        \midrule
        {question generation} & 13 & 38$k$ \\ 
        {answer generation} &  16 & 53$k$ \\ 
        {classification}   & 12 & 36$k$ \\ 
        {incorrect answer generation} & 8 & 18$k$ \\ 
        {minimal modification}  & 10 & 39$k$ \\ 
        {verification}  & 2 & 9$k$ \\ 
        \midrule
        Total & 61 & 193$k$ \\
        \bottomrule
    \end{tabular}
    \caption{Task categories and their statistics. 
    }
    \label{tab:taskcategories}
\end{table}

In total, there are 61 tasks, which are categorized into 6 semantic categories (Table~\ref{tab:taskcategories}). 
We assigned these broad categories to the tasks to understand their collective behavior in the experiments. 
It is noteworthy that, despite the apparent resemblance of the tasks included in the same category,  
any pair of tasks are distinct. 
For example, while \emph{question generation} is part of Quoref, CosmosQA, and QASC, each has its own separate variant of the question generation task (see Fig.\ref{fig:task_specification} in Appendix). 

\subsubsection{Mapping Raw Instructions to Schema } 
\label{subsec:maptoschema}
     We manually fill in the fields of our instruction schema  with the content from the crowdsourcing instructions.
    For instance, parts of the raw instructions that are highlighted for emphasis are incorporated as part of our \emph{emphasis/caution} field. 
The modifications suggested in this step were applied by one author and were verified by another author.\footnote{On average, the process of data curation for each task takes around 5 hrs-34 hrs (details in Appendix; Table~\ref{tab:datacuration}).} 

\vspace{-.2cm}
\paragraph{Improving description quality and consistency.}
    We edit raw instructions to ensure their quality. Particularly, we fix  writing issues (typos, ambiguities, etc.) and  redact repetitions. 
    While repetition often helps in augmenting human understanding, short and concise instructions are often more effective for computers due to their limited attention span~\cite{beltagy2020longformer}. 
    
\vspace{-.2cm}
\paragraph{Augmenting examples and reasons.}
 There is a large variance in the number of examples provided in the raw instructions. Instructions often include more positive examples, or some instructions do not include any negative examples (e.g., QASC). 
Whenever possible, we add negative examples such that each task has at least two negative examples. 
Furthermore, not  all raw instructions contain \textsc{reasons} or \textsc{suggestions} for each of their examples. For example, positive examples are usually not accompanied by explanations, and most datasets do not include suggestions. 
We add them, wherever such information is missing in the instructions. 

\vspace{-.2cm}
\paragraph{Collecting input/output instances for subtasks.} 
Most of our tasks are the intermediate steps in the crowdsourcing process. 
    Therefore, to extract input/output instances for each task, we need to parse the raw annotations of crowdworkers for every step. Since each dataset stores its annotations in a slightly different format, extracting and unifying such intermediate annotations can be non-trivial. 
\vspace{-.2cm}
\paragraph{Verification.} 
\changed{
An annotator verified the quality of the resulting data in consultation with dataset authors. 
The annotator iterated on the authors’ feedback (avg of 3 iters) until they were 
satisfied. 
}

\vspace{-.2cm}
\paragraph{Quality assessment.}
We ask independent human annotators to answer 240 random instances (20 instances from 12 random tasks, used later for our evaluation~\S\ref{subsec:split}). 
The subsequent evaluation of the human-generated responses results in more than 96\% accuracy, which indicates that humans can effortlessly understand and execute our instructions. 



\subsubsection{\name\ Statistics}
\label{subsec:dataset:statistics}

In summary, \name\ consists of  subtasks  each with a set of instructions and input/output instances (Fig.\ref{fig:examples} and \ref{fig:schema_plate}). The complete list of instructions is included in the appendix. In total, the dataset includes 61 tasks and 193$k$ instances.
Table~\ref{tab:taskcategories} shows data statistics for each task category.\footnote{We limit the number of instances in each task to $6.5k$ to avoid massive instance imbalance.} On average, instructions  contain 4.9 positive examples and 2.2 negative examples. 
The longest element of instructions is usually \textsc{Definitions} with 65.5 tokens and the shortest is \textsc{title} with 8.3 tokens (more statistics in Table~\ref{tab:schemastat}).

\begin{table}[ht]
    \centering
    \small
    \begin{tabular}{lc}
        \toprule
        statistic & value  \\ 
        \midrule
        ``title'' length  & 8.3 tokens \\ 
        ``prompt'' length  & 12.6 tokens \\ 
        ``definition'' length & 65.5 tokens \\ 
        ``things to avoid'' length  & 24.1 tokens\\ 
        ``emphasis/caution'' length  & 45.0 tokens\\
        ``reason'' length  & 24.9 tokens\\ 
        ``suggestion'' length & 19.6 tokens\\ 
        num of positive examples & 4.9 \\ 
        num of negative examples & 2.2 \\ 
        \bottomrule
    \end{tabular}
    \caption{
    Statistics of \name{}
}
    \label{tab:schemastat}
\end{table}

\section{Problem Setup and Models }
\label{subsec:setup}
\changed{
Here we define different cross-task generalization settings  (\S \ref{subsec:split}) and the models (\S\ref{subsec:models}). 
}

\subsection{Task Splits and Generalizations Types}
\label{subsec:split}


\paragraph{Random split.}
This setup follows the common practice in benchmarking NLP models with random data splits. Here, two tasks from each task category (Table~\ref{tab:taskcategories}) in \name{} are randomly selected for evaluation, and the rest of the tasks are used for training. This leads to 12 tasks in \task{unseen} and 49 tasks in \task{seen}.\footnote{Those tasks that do not accept a relatively reliable automatic evaluation are excluded from \task{unseen}. }

\paragraph{Leave-one-out generalization.}
To better understand the nature of cross-task generalization, we study more restrictive settings of dividing training and evaluation tasks. 

\noindent \ul{leave-one-category}: evaluates how well a model  generalizes to a task category if it is trained on others  -- no task of that category is in \task{seen}.

\noindent \ul{leave-one-dataset}: evaluates how well a model can generalize to all tasks in a particular dataset  if it is trained on all other tasks -- no task of that dataset is in \task{seen}.
This split prevents any leakage across tasks that belong to the same source datasets. 

\noindent \underline{leave-one-task}: evaluates how well a model can learn a single task by training on all other tasks. \\

\subsection{Models}
\label{subsec:models}
We build models using pre-trained LMs with encoder-decoder architectures BART~\cite{lewis2019bart} for fine-tuning and GPT3~\cite{brown2020language} for few-shot experiments. 
 \paragraph{Encoding instructions and instances.} 
  For every problem setup, we map a given instruction $I_t$ and an input instance $x$ into a textual format and decode an output $y$ and obtain $enc(I_t, x)$. 
This encoding function is then fed to an encoder-decoder model to predict $y$: $M:enc(I_t, x) \rightarrow y$.

\begin{figure}
\centering
\begin{boxedminipage}{\columnwidth}
\begin{equation*} 
    \small
    \begin{split}
        & \mathtt{\small Prompt:} \;  \I{t}{prompt}  \\ 
        & \mathtt{\small Definition:} \;  \I{t}{Definition}  \\  
        & \mathtt{\small Things \; to \; Avoid:} \;  \I{t}{avoid.}  \\  
        & \mathtt{\small Emphasis \& Caution:} \;  \I{t}{emph.}  \\  
        & \textnormal{}\mathtt{\small Negative Example1-}   \\ 
        & \hspace{0.7cm} \mathtt{\small input:} \; \I{t}{pos. ex.}, \mathtt{\small output:} \; \I{t}{pos. ex.},
         \mathtt{\small reason:} \; \I{t}{pos. ex.}  \\  
        & \mathtt{\small Positive Example1-}   \\ 
        & \hspace{0.7cm} \mathtt{\small input:} \; \I{t}{pos. ex.}, 
          \mathtt{\small output:} \; \I{t}{pos. ex.} \mathtt{\small reason:} \; \I{t}{pos. ex. }  \\  
        & \mathtt{\small input:} \; x,  \mathtt{\small output:}  \textnormal{''}
    \end{split}
\end{equation*}
\end{boxedminipage}
\caption{Encoding instruction $I_t$, where $I_t^c$ refers to the text of a component $c$ in the instruction schema.}
\label{fig:encoding}
\end{figure}
Encoding instances follows a standard NLP paradigm of mapping an input instance to text. 
Each instruction $I_t$ consists of multiple elements as described in our instruction schema (\S\ref{subsec:schema}). Here, we map each element of the instruction to a textual format and append it before the input instance.  Fig.\ref{fig:encoding} shows how we encode the full instruction. 

To study the impact of each instruction element for cross-task generalization, we compare these encodings: (1) \textsc{prompt}, (2) \textsc{pos. examples}, (3) \textsc{prompt + definition},  (4) \textsc{prompt + things to avoid},  (5) \textsc{prompt + emphasis} , (6) \textsc{prompt + pos. examples}, (7) \textsc{prompt + + definition + pos. examples}, and (8) \textsc{Full instruction}. 
\changed{
 Each of these (e.g., \textsc{prompt} and \textsc{pos. examples}) correspond to prompting setups in the recent literature~\cite{scao2021many,lu2021fantastically}. 
} 

 \begin{table*}[t]
    \small 
    \centering
        \begin{tabular}{clcccc}
        \toprule
         \makecell{model ↓} & \makecell{evaluation set \task{unseen} →}    & \makecell{random split\\of tasks}   &\makecell{leave-one-\\category (QG)} & \makecell{leave-one-\\dataset (QASC)} & \makecell{leave-one-\\task (QASC QG)} \\
        \cmidrule(lr){1-1} \cmidrule(lr){2-2} \cmidrule(lr){3-3} \cmidrule(lr){4-4} \cmidrule(lr){5-5} \cmidrule(lr){6-6} 
        \multirow{2}{*}{\makecell{BART (fine-Tuned)} } & \textsc{No instructions} & 13 & 6 & 37 & 20 \\
        & \textsc{Full instructions} & \textbf{32} & \textbf{17} & \textbf{51} & \textbf{56} \\
        \midrule
        GPT3 (not fine-tuned)& \textsc{Full instructions} & 24 & 33  & 22 & 33 \\
        \bottomrule
        \end{tabular}
    \caption{Cross-task generalization of BART under various splits (\S\ref{subsec:split}).
    Fine-tuned BART shows improved performance when provided with instructions. 
     It also archives better performance than GPT3, despite being over $1k$ times smaller. 
    \changed{All numbers are ROUGE-L. }
    }
    \label{tab:bart:generalization:all:splits}
\end{table*}

\vspace{-.2cm}
\paragraph{BART.}
\label{sec:bart}
We use  BART (base)~\cite{lewis2019bart} which allows us to fine-tune its  model parameters. 
This is an encoder-decoder architecture with $140m$ parameters.
For each setup, the input is encoded using different instruction elements,  trained  on all \task{seen} tasks, and evaluated  on \task{unseen} (\S\ref{subsec:split}). 

\vspace{-.2cm}
\paragraph{GPT3.}
As a comparison, we evaluate
GPT3~\cite{brown2020language} which is a $175B$ parameter autoregressive LM ($\times1.2k$ larger than BART) and has shown promising results in mimicking demonstrations provided in its prompt.
We cannot fine-tune the parameters of this massive model and use it as-is  
under its default setting on the evaluation tasks in \task{unseen} (\S\ref{subsec:split}) using the  encoding introduced earlier.

\begin{table*}[t]
    \small 
    \centering
\begin{tabular}{llcccccc|l}
\toprule
model ↓  & task category → & QG & AG & CF & IAG & MM & VF & avg  \\
\midrule
\multirow{9}{*}{\makecell{BART\\(fine-tuned)}} &  \textsc{No Instruction} & 26 & 6 & 0 & 21 & 33 & 7 & 13  \\
\cmidrule(lr){2-9}
& \textsc{prompt} & 27 & 22 & 7 & 22 & 34 & \textbf{9} & 20  \\
& {\ \ \ +\textsc{definition}} & 35 & 24 & 50 & 25 & 36 & 7 & 30$\uparrow$ (+50) \\
& { \ \ \ +\textsc{things to avoid}} & 33 & 24 & 4 & 24 & \textbf{58} & \textbf{9} & 25$\uparrow$ (+25) \\
&  {\ \ \ +\textsc{emphasis}} & 38 & 23 & 16 & \textbf{26} & 49 & 3 & 26$\uparrow$ (+30) \\
&  {\ \ \ +\textsc{pos. examples}} & 53 & 22 & 14 & 25 & 17 & 7 & 23$\uparrow$ (+15)  \\
&  {\ \ \ +\textsc{definition+pos. examples}} & 51 & 23 & \textbf{56} & 25 & 37 & 6 & 33$\uparrow$ (+65)  \\
&  \textsc{pos. examp.} & \textbf{55} & 6 & 18 & 25 & 8 & 6 & 20  \\
&  \textsc{Full Instruction} & 46 & \textbf{25} & 52 & 25 & 35 & 7 & 32$\uparrow$ (+60) \\
\midrule 
\makecell{GPT3\\(not fine-tuned)} & \textsc{Full Instruction} & 33 & 18 & 8 & 12 & 60 & 11 & 24 (+11)  \\
\bottomrule
\end{tabular}
 \caption{
    Cross-task generalization 
    under random split (\S\ref{subsec:split}).
    Models show improved results when provided with instructions. 
    \changed{
        The numbers in parenthesis indicate absolute gains compared to `\textsc{No Instructions}' baseline. 
    }
    Fine-tuned BART archives better performance than GPT3, despite being over $1k$ times smaller. 
    Category names: QG: Question Generation, AG: Answer Generation, CF: Classification, IAG: Incorrect Answer Generation, MM: Minimal Text Modification, VF: Verification.
    \changed{All numbers are ROUGE-L (in percentage). }
    }
    \label{tab:random:splitfull2}
\end{table*}

\section{Experiments}
\label{sec:experiments}
\vspace{-.1cm}
\paragraph{Evaluation metrics.}
We treat all of our tasks as text generation problems and evaluate them with 
automated evaluation metrics for text generation. 
In particular, we use 
ROUGE-L~\cite{lin2004rouge} to automatically evaluate the generated outputs.\footnote{
Our experiments show that other metrics, e.g. BLEURT~\cite{sellam2020bleurt} are also correlated with ROUGE-L, which has also been used in generative QA tasks.
}

\vspace{-.2cm}
\paragraph{Implementation details.}
For BART, our  models are trained for 3 epochs with a learning rate of 5e-5 for a given training split and input encoding. For GPT3, we use the {\texttt{davinci-instruct}} engine and produce outputs with greedy decoding, 
generating up to a maximum number of tokens of 16 (the default value). We use the default stop condition which is 2 newline tokens.\footnote{The relevant code is available at:  
\url{https://github.com/allenai/natural-instructions-v1}
}

\subsection{Generalization Under Various Task Splits}
\label{sec:gen:various:splits}

\changed{
Table~\ref{tab:bart:generalization:all:splits} reports the results of the BART model train and evaluated with various task splits (\S\ref{subsec:split})}. 
For comparison, we evaluate GPT3 which uses no fine-tuning, unlike BART that is fine-tuned with the \task{seen} tasks.
The first column corresponds to random split of tasks, while the remaining columns report cross-task generalization results of the BART model under 
\changed{
leave-one-$x$
}
splits (\S\ref{subsec:split}). 
For 
\changed{
$x =$ \ul{category},}
the tasks in \emph{question-generation} category are held out during training. 
For 
\changed{
$x =$ \ul{dataset},}
the tasks that were extracted from the \emph{QASC} dataset were excluded from training. 
For 
\changed{
$x =$ \ul{task},}
we train a model on all tasks, except \emph{QASC question generation} task which is used for evaluation.

\vspace{-.2cm}
\paragraph{Instructions benefit cross-task generalization.} 
The results indicate that BART benefits from instructions in generalizing to new tasks, regardless of task splits.  
For example, under random split, the model using \textsc{Full Instructions}  results in +19\% gains over a model that is not using instructions.  
 This is particularly interesting for 
leave-one-\ul{category}-out split
since the trained model can generalize to the tasks of a particular semantic category, without being exposed to it. 
In comparison to GPT3, the fine-tuned BART model that utilizes instructions  achieves a stronger performance despite being $\times 1k$ smaller than GPT3. 
For example, a BART models using \textsc{Full Instructions} achieves 8\% higher performance than GPT3 under random split of tasks.

Note that the absolute values in leave-one-category are lower due to the difficulty of this setup compared to, for example, the random split setup. 
While all settings involve evaluating on tasks not seen during training, the leave-one-category setting enforces more dissimilarity among training and evaluation tasks.

\subsection{Generalization Under  Instruction Encoding and Task Categories}
Table~\ref{tab:random:splitfull2} reports the results of the BART model  
per encodings of different instruction elements (\S\ref{subsec:models}) and for different task categories.
The table shows that encoding more elements of the instructions generally achieves better results than just using  \textsc{prompt} or \textsc{positive examples}. 
It additionally shows that the benefit of the instruction elements seems to depend on the target task category. 
We observe that the \emph{question-generation} (QG) tasks benefit the most from \textsc{positive examples}, whereas  in \emph{classification} (CF),
\textsc{positive examples} are of little help. We hypothesis this is because it is easier to mimic question-generation based on a few examples, whereas it is difficult to define classes via a few examples, where \textsc{definition} can be more helpful. 
The models show little improvement in \emph{verification} (VF). 
We hypothesize these tasks are inherently more difficult,  partially because of  their distinctness from the rest of the tasks in the dataset. 
We hope future work on this line will study a wider variety of tasks and will improve our understanding of such failure cases. 

\subsection{Generalization vs. Number of Seen Tasks}
\label{subsec:supervision:size:experiment}
Fig.\ref{fig:scaling:tasks} compares the impact of the number of seen tasks for cross-task generalization. 
For supervision, we randomly sample a few tasks as \task{seen} and evaluate on 6 tasks (one from each category). 
(each point in the figure is averaged over 5 random subsamples.) 
The results
show that with \textsc{no-instruction} encoding there is no tangible value in observing more tasks.  
In contrast, the generalization of the models that encode instructions improves with observing more tasks. 
This is an exciting observation since it suggests that scaling up our dataset to more tasks may lead to stronger instruction-following systems.

\subsection{Analyses} \label{subsec:task-specific}

\paragraph{Upperbound: Task-specific Models.}
For each task, we obtain a task-specific model (\S~\ref{subsec:input:output}) by training BART separately on each task's annotated training data. We evaluate these task-specific models to obtain a loose estimate of \emph{upperbounds} for each task. 
On average, task-specific models score 66\% which is considerably higher than our models' best generalization (32\%; Table~\ref{tab:bart:generalization:all:splits}).
This indicates that { there is considerable room for improving generalization-based models} that use instructions. 

\begin{table}
    \centering
    \small
    \resizebox{0.99\linewidth}{!}{
    \begin{tabular}{llcc}
        \toprule
        Model ↓ & Split ↓ & \makecell{w/ neg.\\examples}  & \makecell{w/o neg.\\examples} \\
        \midrule
        \multirow{5}{*}{BART} & random & 32 & {\bf 35} \\ 
        & leave-one-$x$ \\ 
        & \ $\drsh x=$ category (AG) & 19 & {\bf 21} \\ 
        & \ $\drsh x=$ dataset (Quoref) & 37 & 37  \\ 
        & \ $\drsh x=$ task (QASC QG) & 56 &  {\bf 57} \\ 
        \midrule
        GPT3 & - & 24 & {\bf 44} \\ 
        \bottomrule
    \end{tabular}
    }
    \caption{
        Effect of excluding negative examples from \textsc{Full Instruction} encoding. Negative instructions are surprisingly difficult for the models to learn from. 
    }
    \label{tab:negative:examples}
\end{table}

\paragraph{Impact of Negative Examples.}
Crowdsourcing instructions often include negative examples to exemplify undesirable responses. 
We study how negative examples in instructions affect cross-task generalization. 
Our cases study (Table~\ref{tab:negative:examples}) indicates that the models work better \emph{without} (w/o) negative examples,  
contrary to the previously-observed benefits of other instructional elements (e.g., definition, positive examples). 
This is aligned with the previous studies ~\cite{xuan2020hard,lin2003bootstrapped} that discuss the challenges of learning from negative examples.
Interestingly, GPT3's drop (44 vs 24) is more significant than BART (35 vs 32), showing that BART can partly recover through the training step. 

\begin{table*}[ht]
    \centering
    \resizebox{0.78\textwidth}{!}{
    \footnotesize
    \begin{tabular}{p{4.5cm}p{3.5cm}p{6.5cm}}
        \toprule
        Category & Helpful Fields & Explanation \\
        \midrule
        Question Generation (QG) & 1. \textsc{Definition} & - Provides a holistic picture of the task.\\
        & 2. \textsc{Emphasis \& Caution} & - Provides key information for solving the task.\\
        & 3. \textsc{Positive Examples} & - This gives an idea of what is expected in the output.\\
        & 4. \textsc{Negative Examples} & - Good to know the common mistakes people do.\\ 
        \midrule
        Answer Generation (AG) & \textsc{1. Prompt} & - It limits the exploration space to question spans.\\
        & \textsc{2. Definition} & - Provides a general understanding of the task. \\
        & \textsc{3. Positive Examples} & - Reason field is very helpful.\\
        \midrule
        Classification (CF) & \textsc{1. Definition} & - The task is unclear without this field.\\
        \midrule
        Incorrect Answer Generation (IAG) & \textsc{1. Definition} & - Helps understand the utility of such a task.\\
        & \textsc{2. Emphasis \& Caution} & - Source of some useful shortcuts.\\
        & \textsc{3. Positive Examples} & - Helps in understanding the type of questions asked.\\
        \midrule
        Minimal Text Modification (MM) & \textsc{1. Things to Avoid} & - Provides critical information.\\
        \midrule
        Verification (VF) & \textsc{1. Definition} & - Makes the task easy to understand.\\
        & \textsc{2. Things to avoid} & - Contains useful tips required for this task.\\
        & \textsc{3. Positive Examples} & - Exemplifies task understanding.\\ 
        & \textsc{4. Negative examples} & - Helps avoid potential mistakes.\\
        \bottomrule
        \end{tabular}
        }
        \caption{Results of humans' perceived importance of instruction elements. Our annotators, for example, find \textsc{Definition} and \textsc{Thing to Avoid} to be helpful for \textit{Classification} and \textit{Minimal Text Modification} tasks, respectively.}
        \label{Tab:User:Study}
\end{table*}

\paragraph{Error Analysis.}
We randomly sample 30 erroneous predictions of our fine-tuned BART on 3 distinct tasks (Winogrande answer generation; QASC question generation; MC-TACO incorrect answer generation). We categorize the errors into common patterns (Table~\ref{Tab: Error Analysis}).
\begin{table}[ht]
    \centering
    \small
    \begin{tabular}{lc}
        \toprule
         error type & BART  \\
      \midrule
            {\color{brown} \textit{Generates a nonsensical/vague question}} &  47\\
            {\color{brown}\textit{Generate an invalid question}} &  8\\
            {\color{brown}\textit{Generates a yes/no question}} &  4\\
            {\color{brown}\textit{Copies the given fact or a subset of it}} &  3 \\
            {\color{brown}\textit{Generates unanswerable questions}}  &3\\
        \bottomrule
        \end{tabular}
        \caption{
            Percentage of  errors on QASC QG task. 
            The numbers do not sum to 100 since the error types are not mutually exclusive.
        }
        \label{Tab: Error Analysis}
\end{table}

Unlike GPT3 which generally suffers from generating redundant content irrelevant to instructions, our BART model provides more control, however, it can fail to generate proper output. 
Here are several erroneous predictions from our model: 


\egbox{
\fontsize{8pt}{10pt}\selectfont
    {\color{brown}\textit{Generates a nonsensical/vague question}}\\
    \textbf{BART input:}[...]input: Fact: graduations are used for measuring the volume of substances. output: ?\\
    \textbf{BART output:} How many substances are there in volume?
}

\egbox{
\fontsize{8pt}{10pt}\selectfont
    {\color{brown}\textit{Copies the given fact or a subset of it}}\\
    \textbf{BART input:}[...] input: Fact: If someone is not breathing, begin with ventilation. output: ? \\
    \textbf{BART output:} If someone is breathing, begin with ventilation.
}


\egbox{
\fontsize{8pt}{10pt}\selectfont
    {\color{brown}\textit{Generates a yes/no question }}\\
    \textbf{BART input:}[...] input: Fact: Lack of saliva hinders the break down of food into components the body can absorb. output: ?
    \textbf{BART output:} Does saliva hinders the break down of food into components the body can absorb?
}

\paragraph{Perceived Impact of Instruction Elements.}
We survey human  annotators to find out the value of instruction elements to humans.
Except for the negative examples which were shown to be difficult for models, we observe similar trends between  humans' perceived value of those elements (Table~\ref{Tab:User:Study}) and their contributions to the model performance (Table~\ref{tab:random:splitfull2}).  
For example, humans viewed  \textsc{Definition} and \textsc{Things to Avoid} as necessary fields for  \emph{classification}  and \emph{minimal text modification} categories, respectively, which is compatible with our empirical observations (e.g., \textsc{prompt + definition} has the highest score on CF category in Table~\ref{tab:random:splitfull2}).

\section{Conclusion}
\label{sec:discussion}
In this paper, we studied the goal of building models that  generalize to new tasks by encoding and understanding crowdsourcing instructions. 
We introduced \name{}, which is built based on existing crowdsourced datasets,  that enables building such models and systematically evaluate them. 
To the best of our knowledge, this is the first work to show the benefit of instructions towards improved cross-task generalization. 
Additionally, we observe that our proposed task has a large room for improvement, which we believe 
will bring more attention to building stronger models that can generalize to a wider range of tasks.

\section*{Acknowledgements}
We thank OpenAI for providing access to the GPT3 API, 
authors who generously shared their dataset templates with us, Matt Peters and Nicholas Lourie for helpful input, the Beaker team for their support with experiments, and the anonymous reviewers for their helpful feedback. 
The support of DARPA SAIL-ON,  DARPA CHESS program,  NSF IIS-2044660, ONR N00014-18-1-2826,
and Paul G. Allen Foundation is gratefully acknowledged.

\bibliography{ref}

\begin{thebibliography}{39}
\expandafter\ifx\csname natexlab\endcsname\relax\def\natexlab#1{#1}\fi

\bibitem[{Aghajanyan et~al.(2021)Aghajanyan, Gupta, Shrivastava, Chen,
  Zettlemoyer, and Gupta}]{aghajanyan2021muppet}
Armen Aghajanyan, Anchit Gupta, Akshat Shrivastava, Xilun Chen, Luke
  Zettlemoyer, and Sonal Gupta. 2021.
\newblock Muppet: Massive multi-task representations with pre-finetuning.
\newblock In \emph{Proceedings of EMNLP}, pages 5799--5811.

\bibitem[{Ali(1981)}]{ali1981use}
Ali~M Ali. 1981.
\newblock The use of positive and negative examples during instruction.
\newblock \emph{Journal of instructional development}, 5(1):2--7.

\bibitem[{Beltagy et~al.(2020)Beltagy, Peters, and
  Cohan}]{beltagy2020longformer}
Iz~Beltagy, Matthew~E Peters, and Arman Cohan. 2020.
\newblock Longformer: The long-document transformer.
\newblock \emph{arXiv preprint arXiv:2004.05150}.

\bibitem[{Brown et~al.(2020)Brown, Mann, Ryder, Subbiah, Kaplan, Dhariwal,
  Neelakantan, Shyam, Sastry, Askell, Agarwal, Herbert-Voss, Krueger, Henighan,
  Child, Ramesh, Ziegler, Wu, Winter, Hesse, Chen, Sigler, Litwin, Gray, Chess,
  Clark, Berner, McCandlish, Radford, Sutskever, and
  Amodei}]{brown2020language}
Tom Brown, Benjamin Mann, Nick Ryder, Melanie Subbiah, Jared~D Kaplan, Prafulla
  Dhariwal, Arvind Neelakantan, Pranav Shyam, Girish Sastry, Amanda Askell,
  Sandhini Agarwal, Ariel Herbert-Voss, Gretchen Krueger, Tom Henighan, Rewon
  Child, Aditya Ramesh, Daniel Ziegler, Jeffrey Wu, Clemens Winter, Chris
  Hesse, Mark Chen, Eric Sigler, Mateusz Litwin, Scott Gray, Benjamin Chess,
  Jack Clark, Christopher Berner, Sam McCandlish, Alec Radford, Ilya Sutskever,
  and Dario Amodei. 2020.
\newblock Language models are few-shot learners.
\newblock In \emph{NeurIPS}, volume~33, pages 1877--1901. Curran Associates,
  Inc.

\bibitem[{Caruana(1997)}]{caruana1997multitask}
Rich Caruana. 1997.
\newblock Multitask learning.
\newblock \emph{Machine learning}, 28(1):41--75.

\bibitem[{Dasigi et~al.(2019)Dasigi, Liu, Marasovic, Smith, and
  Gardner}]{dasigi2019quoref}
Pradeep Dasigi, Nelson~F Liu, Ana Marasovic, Noah~A Smith, and Matt Gardner.
  2019.
\newblock Quoref: A reading comprehension dataset with questions requiring
  coreferential reasoning.
\newblock In \emph{Proceedings of EMNLP-IJCNLP}, pages 5927--5934.

\bibitem[{Dua et~al.(2019)Dua, Wang, Dasigi, Stanovsky, Singh, and
  Gardner}]{dua2019drop}
Dheeru Dua, Yizhong Wang, Pradeep Dasigi, Gabriel Stanovsky, Sameer Singh, and
  Matt Gardner. 2019.
\newblock Drop: A reading comprehension benchmark requiring discrete reasoning
  over paragraphs.
\newblock In \emph{Proceedings of NAACL}, pages 2368--2378.

\bibitem[{Efrat and Levy(2020)}]{efrat2020turking}
Avia Efrat and Omer Levy. 2020.
\newblock The turking test: Can language models understand instructions?
\newblock \emph{arXiv preprint arXiv:2010.11982}.

\bibitem[{Gupta et~al.(2021)Gupta, Kamath, Kembhavi, and
  Hoiem}]{Gupta2021TowardsGP}
Tanmay Gupta, A.~Kamath, Aniruddha Kembhavi, and Derek Hoiem. 2021.
\newblock Towards general purpose vision systems.
\newblock \emph{ArXiv}, abs/2104.00743.

\bibitem[{Hase and Bansal(2021)}]{hase2021can}
Peter Hase and Mohit Bansal. 2021.
\newblock When can models learn from explanations? a formal framework for
  understanding the roles of explanation data.
\newblock \emph{arXiv preprint arXiv:2102.02201}.

\bibitem[{Huang et~al.(2019)Huang, Le~Bras, Bhagavatula, and
  Choi}]{huang2019cosmos}
Lifu Huang, Ronan Le~Bras, Chandra Bhagavatula, and Yejin Choi. 2019.
\newblock Cosmos qa: Machine reading comprehension with contextual commonsense
  reasoning.
\newblock In \emph{Proceedings of EMNLP-IJCNLP}, pages 2391--2401.

\bibitem[{Khashabi et~al.(2018)Khashabi, Chaturvedi, Roth, Upadhyay, and
  Roth}]{khashabi2018looking}
Daniel Khashabi, Snigdha Chaturvedi, Michael Roth, Shyam Upadhyay, and Dan
  Roth. 2018.
\newblock Looking beyond the surface: A challenge set for reading comprehension
  over multiple sentences.
\newblock In \emph{Proceedings of NAACL}, pages 252--262.

\bibitem[{Khashabi et~al.(2017)Khashabi, Khot, Sabharwal, and
  Roth}]{khashabi2017learning}
Daniel Khashabi, Tushar Khot, Ashish Sabharwal, and Dan Roth. 2017.
\newblock Learning what is essential in questions.
\newblock In \emph{Proceedings of CoNLL}, pages 80--89.

\bibitem[{Khashabi et~al.(2020)Khashabi, Min, Khot, Sabharwal, Tafjord, Clark,
  and Hajishirzi}]{khashabi2020unifiedqa}
Daniel Khashabi, Sewon Min, Tushar Khot, Ashish Sabharwal, Oyvind Tafjord,
  Peter Clark, and Hannaneh Hajishirzi. 2020.
\newblock {UnifiedQA:} crossing format boundaries with a single qa system.
\newblock In \emph{Proceedings of EMNLP: Findings}, pages 1896--1907.

\bibitem[{Khot et~al.(2020)Khot, Clark, Guerquin, Jansen, and
  Sabharwal}]{khot2020qasc}
Tushar Khot, Peter Clark, Michal Guerquin, Peter Jansen, and Ashish Sabharwal.
  2020.
\newblock {QASC: A dataset for question answering via sentence composition}.
\newblock In \emph{Proceedings of AAAI}.

\bibitem[{Le~Scao and Rush(2021)}]{scao2021many}
Teven Le~Scao and Alexander~M Rush. 2021.
\newblock How many data points is a prompt worth?
\newblock In \emph{Proceedings of NAACL-HLT}, pages 2627--2636.

\bibitem[{Lewis et~al.(2019)Lewis, Liu, Goyal, Ghazvininejad, Mohamed, Levy,
  Stoyanov, and Zettlemoyer}]{lewis2019bart}
Mike Lewis, Yinhan Liu, Naman Goyal, Marjan Ghazvininejad, Abdelrahman Mohamed,
  Omer Levy, Ves Stoyanov, and Luke Zettlemoyer. 2019.
\newblock {BART}: Denoising sequence-to-sequence pre-training for natural
  language generation, translation, and comprehension.
\newblock In \emph{Proceedings of ACL}.

\bibitem[{Lin(2004)}]{lin2004rouge}
Chin-Yew Lin. 2004.
\newblock Rouge: A package for automatic evaluation of summaries.
\newblock In \emph{Text summarization branches out}, pages 74--81.

\bibitem[{Lin et~al.(2019)Lin, Tafjord, Clark, and Gardner}]{lin2019reasoning}
Kevin Lin, Oyvind Tafjord, Peter Clark, and Matt Gardner. 2019.
\newblock Reasoning over paragraph effects in situations.
\newblock In \emph{Proceedings of the 2nd Workshop on Machine Reading for
  Question Answering}, pages 58--62.

\bibitem[{Lin et~al.(2003)Lin, Yangarber, and Grishman}]{lin2003bootstrapped}
Winston Lin, Roman Yangarber, and Ralph Grishman. 2003.
\newblock Bootstrapped learning of semantic classes from positive and negative
  examples.
\newblock In \emph{Proceedings of ICML Workshop on The Continuum from Labeled
  to Unlabeled Data}, volume~1, page~21.

\bibitem[{Liu et~al.(2021)Liu, Yuan, Fu, Jiang, Hayashi, and
  Neubig}]{liu2021pre}
Pengfei Liu, Weizhe Yuan, Jinlan Fu, Zhengbao Jiang, Hiroaki Hayashi, and
  Graham Neubig. 2021.
\newblock Pre-train, prompt, and predict: A systematic survey of prompting
  methods in natural language processing.
\newblock \emph{arXiv preprint arXiv:2107.13586}.

\bibitem[{Lu et~al.(2021)Lu, Bartolo, Moore, Riedel, and
  Stenetorp}]{lu2021fantastically}
Yao Lu, Max Bartolo, Alastair Moore, Sebastian Riedel, and Pontus Stenetorp.
  2021.
\newblock Fantastically ordered prompts and where to find them: Overcoming
  few-shot prompt order sensitivity.
\newblock \emph{arXiv preprint arXiv:2104.08786}.

\bibitem[{McCann et~al.(2018)McCann, Keskar, Xiong, and
  Socher}]{mccann2018natural}
Bryan McCann, Nitish~Shirish Keskar, Caiming Xiong, and Richard Socher. 2018.
\newblock The natural language decathlon: Multitask learning as question
  answering.

\bibitem[{Mishra et~al.(2020)Mishra, Mitra, Varshney, Sachdeva, and
  Baral}]{mishra2020towards}
Swaroop Mishra, Arindam Mitra, Neeraj Varshney, Bhavdeep Sachdeva, and Chitta
  Baral. 2020.
\newblock Towards question format independent numerical reasoning: A set of
  prerequisite tasks.
\newblock \emph{arXiv preprint arXiv:2005.08516}.

\bibitem[{Peters et~al.(2018)Peters, Neumann, Iyyer, Gardner, Clark, Lee, and
  Zettlemoyer}]{peters2018deep}
Matthew~E Peters, Mark Neumann, Mohit Iyyer, Matt Gardner, Christopher Clark,
  Kenton Lee, and Luke Zettlemoyer. 2018.
\newblock Deep contextualized word representations.
\newblock In \emph{Proceedings of NAACL-HLT}, pages 2227--2237.

\bibitem[{Raffel et~al.(2020)Raffel, Shazeer, Roberts, Lee, Narang, Matena,
  Zhou, Li, and Liu}]{raffel2020exploring}
Colin Raffel, Noam Shazeer, Adam Roberts, Katherine Lee, Sharan Narang, Michael
  Matena, Yanqi Zhou, Wei Li, and Peter~J Liu. 2020.
\newblock Exploring the limits of transfer learning with a unified text-to-text
  transformer.
\newblock \emph{Journal of Machine Learning Research}, 21(140):1--67.

\bibitem[{Reynolds and McDonell(2021)}]{reynolds2021prompt}
Laria Reynolds and Kyle McDonell. 2021.
\newblock Prompt programming for large language models: Beyond the few-shot
  paradigm.
\newblock In \emph{Extended Abstracts of the 2021 CHI Conference on Human
  Factors in Computing Systems}, pages 1--7.

\bibitem[{Sakaguchi et~al.(2020)Sakaguchi, Le~Bras, Bhagavatula, and
  Choi}]{sakaguchi2020winogrande}
Keisuke Sakaguchi, Ronan Le~Bras, Chandra Bhagavatula, and Yejin Choi. 2020.
\newblock Winogrande: An adversarial winograd schema challenge at scale.
\newblock In \emph{Proceedings of the AAAI}.

\bibitem[{Sanh et~al.(2022)Sanh, Webson, Raffel, Bach, Sutawika, Alyafeai,
  Chaffin, Stiegler, Raja, Dey, Bari, Xu, Thakker, Sharma, Szczechla, Kim,
  Chhablani, Nayak, Datta, Chang, Jiang, Wang, Manica, Shen, Yong, Pandey,
  Bawden, Wang, Neeraj, Rozen, Sharma, Santilli, Fevry, Fries, Teehan, Scao,
  Biderman, Gao, Wolf, and Rush}]{sanh2021multitask}
Victor Sanh, Albert Webson, Colin Raffel, Stephen Bach, Lintang Sutawika, Zaid
  Alyafeai, Antoine Chaffin, Arnaud Stiegler, Arun Raja, Manan Dey, M~Saiful
  Bari, Canwen Xu, Urmish Thakker, Shanya~Sharma Sharma, Eliza Szczechla,
  Taewoon Kim, Gunjan Chhablani, Nihal Nayak, Debajyoti Datta, Jonathan Chang,
  Mike Tian-Jian Jiang, Han Wang, Matteo Manica, Sheng Shen, Zheng~Xin Yong,
  Harshit Pandey, Rachel Bawden, Thomas Wang, Trishala Neeraj, Jos Rozen,
  Abheesht Sharma, Andrea Santilli, Thibault Fevry, Jason~Alan Fries, Ryan
  Teehan, Teven~Le Scao, Stella Biderman, Leo Gao, Thomas Wolf, and Alexander~M
  Rush. 2022.
\newblock Multitask prompted training enables zero-shot task generalization.
\newblock In \emph{Proceedings of ICLR}.

\bibitem[{Schick and Sch{\"u}tze(2021)}]{schick2020few}
Timo Schick and Hinrich Sch{\"u}tze. 2021.
\newblock Few-shot text generation with natural language instructions.
\newblock In \emph{Proceedings of EMNLP}.

\bibitem[{Sellam et~al.(2020)Sellam, Das, and Parikh}]{sellam2020bleurt}
Thibault Sellam, Dipanjan Das, and Ankur Parikh. 2020.
\newblock Bleurt: Learning robust metrics for text generation.
\newblock In \emph{Proceedings of ACL}, pages 7881--7892.

\bibitem[{Wei et~al.(2022)Wei, Bosma, Zhao, Guu, Yu, Lester, Du, Dai, and
  Le}]{wei2021finetuned}
Jason Wei, Maarten Bosma, Vincent Zhao, Kelvin Guu, Adams~Wei Yu, Brian Lester,
  Nan Du, Andrew~M. Dai, and Quoc~V Le. 2022.
\newblock Finetuned language models are zero-shot learners.
\newblock In \emph{Proceedings of ICLR}.

\bibitem[{Weller et~al.(2020)Weller, Lourie, Gardner, and
  Peters}]{weller-etal-2020-learning}
Orion Weller, Nicholas Lourie, Matt Gardner, and Matthew Peters. 2020.
\newblock Learning from task descriptions.
\newblock In \emph{Proceedings of EMNLP}, pages 1361--1375.

\bibitem[{Xuan et~al.(2020)Xuan, Stylianou, Liu, and Pless}]{xuan2020hard}
Hong Xuan, Abby Stylianou, Xiaotong Liu, and Robert Pless. 2020.
\newblock Hard negative examples are hard, but useful.
\newblock In \emph{Proceedings of ECCV}, pages 126--142. Springer.

\bibitem[{Ye et~al.(2021)Ye, Lin, and Ren}]{ye2021crossfit}
Qinyuan Ye, Bill~Yuchen Lin, and Xiang Ren. 2021.
\newblock Crossfit: A few-shot learning challenge for cross-task generalization
  in nlp.
\newblock In \emph{Proceedings of EMNLP}.

\bibitem[{Ye and Ren(2021)}]{ye2021zero}
Qinyuan Ye and Xiang Ren. 2021.
\newblock Zero-shot learning by generating task-specific adapters.
\newblock \emph{arXiv preprint arXiv:2101.00420}.

\bibitem[{Zhao et~al.(2021)Zhao, Wallace, Feng, Klein, and
  Singh}]{zhao2021calibrate}
Zihao Zhao, Eric Wallace, Shi Feng, Dan Klein, and Sameer Singh. 2021.
\newblock Calibrate before use: Improving few-shot performance of language
  models.
\newblock In \emph{Proceedings of ICML}, pages 12697--12706.

\bibitem[{Zhong et~al.(2021)Zhong, Lee, Zhang, and Klein}]{Zhong2021AdaptingLM}
Ruiqi Zhong, Kristy Lee, Zheng Zhang, and Dan Klein. 2021.
\newblock Adapting language models for zero-shot learning by meta-tuning on
  dataset and prompt collections.
\newblock In \emph{Proceedings of EMNLP: Findings}, pages 2856--2878.

\bibitem[{Zhou et~al.(2019)Zhou, Khashabi, Ning, and Roth}]{zhou2019going}
Ben Zhou, Daniel Khashabi, Qiang Ning, and Dan Roth. 2019.
\newblock “going on a vacation” takes longer than “going for a walk”: A
  study of temporal commonsense understanding.
\newblock In \emph{Proceedings of EMNLP-IJCNLP}, pages 3354--3360.

\end{thebibliography}
\bibliographystyle{acl_natbib}

\clearpage

\appendix

\label{sec:appendix}


\begin{center}
{\Large \textbf{Supplemental Material}}
\end{center}


\section{Datasets and their Templates}
\label{sec:appendix:analysis}

\subsection{Division of Crowdsourcing Instructions into Minimal Tasks}
\label{sec:appendix:division:screenshots}
\begin{figure}
\centering
  \includegraphics[width=0.5\textwidth,trim=0cm 0cm 0cm 0cm]{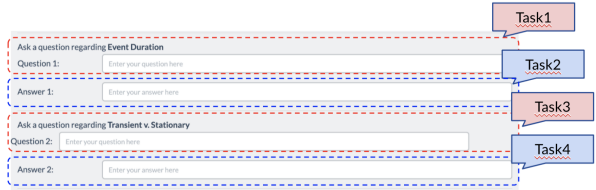}
  \caption{
    Dividing a data creation task into multiple subtasks for the MC-TACO dataset. 
  }
  \label{fig:subtsakdivision}
\end{figure}

Fig.~\ref{fig:subtsakdivision} shows an example of how a task is divided into multiple subtasks for the MC-TACO dataset. MC-TACO has five categories (Event Duration, Event Frequency etc.). Each category contributes to 2 subtasks one for question generation and one for answer generation.

\begin{figure}[H]
\centering
  \includegraphics[scale=0.28,trim=2cm 4cm 0cm 3cm]{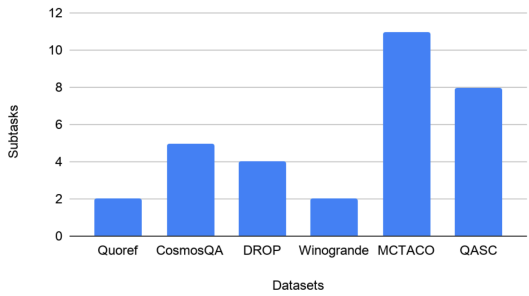}
  \caption{Variations in the number of subtasks}
    \label{fig:no. of subtasks}
\end{figure}

\paragraph{Number of tasks in each dataset.} 
Fig.~\ref{fig:no. of subtasks} illustrates how the number of steps in the data creation process varies across the 6 datasets. QASC and MC-TACO contain a relatively higher number of steps in the data creation process in comparison to DROP, Quoref, CosmosQA, and Winogrande.

\subsection{Analysis of Crowdsourcing Templates}
We analyzed crowdsourcing templates of 6 datasets: CosmosQA~\cite{huang2019cosmos}, 
DROP~\cite{dua2019drop}, 
MC-TACO~\cite{zhou2019going}, 
QASC~\cite{khot2020qasc}, 
Quoref~\cite{dasigi2019quoref}, and
Winogrande~\cite{sakaguchi2020winogrande}. Our intention behind the analysis is to identify similarities and differences across templates and subsequently decide regarding the collection of more templates.
\label{appendix:analysis:templates}

\begin{figure}[H]
    \centering
    \includegraphics[width=0.96\columnwidth]{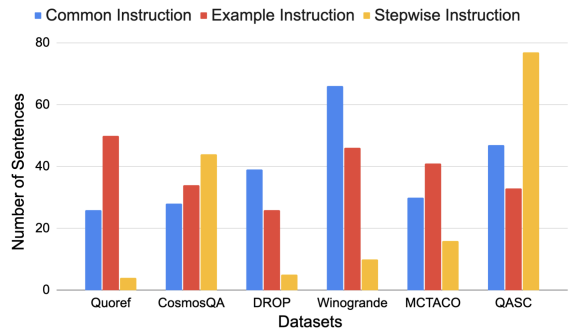}
    \caption{Variation in the number of sentences in the crowdsourcing instructions across datasets}
    \label{fig:size inst}
\end{figure}

\paragraph{Size of the instructions.} We observe significant variation in size across the 6 datasets (Fig.~\ref{fig:size inst}). In the case of QASC, the instruction size associated with each step of the data creation process is very high, whereas for Winogrande, it is exactly the opposite-- instruction size associated with each step of the data creation process is very low. Instead, the size of the common instruction (i.e., the instruction preceding the first step of the data creation process) is high in Winogrande; this is  also seen for DROP. The major mode of instruction varies across datasets. Examples and instructions associated with each step of data creation respectively take up the majority of space in Quoref and CosmosQA.  MC-TACO relies on examples to explain the crowdsourcing task, while Winogrande and QASC depend mostly on common instructions and instructions associated with each step of the data creation process respectively, to explain the task to the crowdworker.

\begin{figure}[H]
\centering
  \includegraphics[width=0.96\columnwidth ]{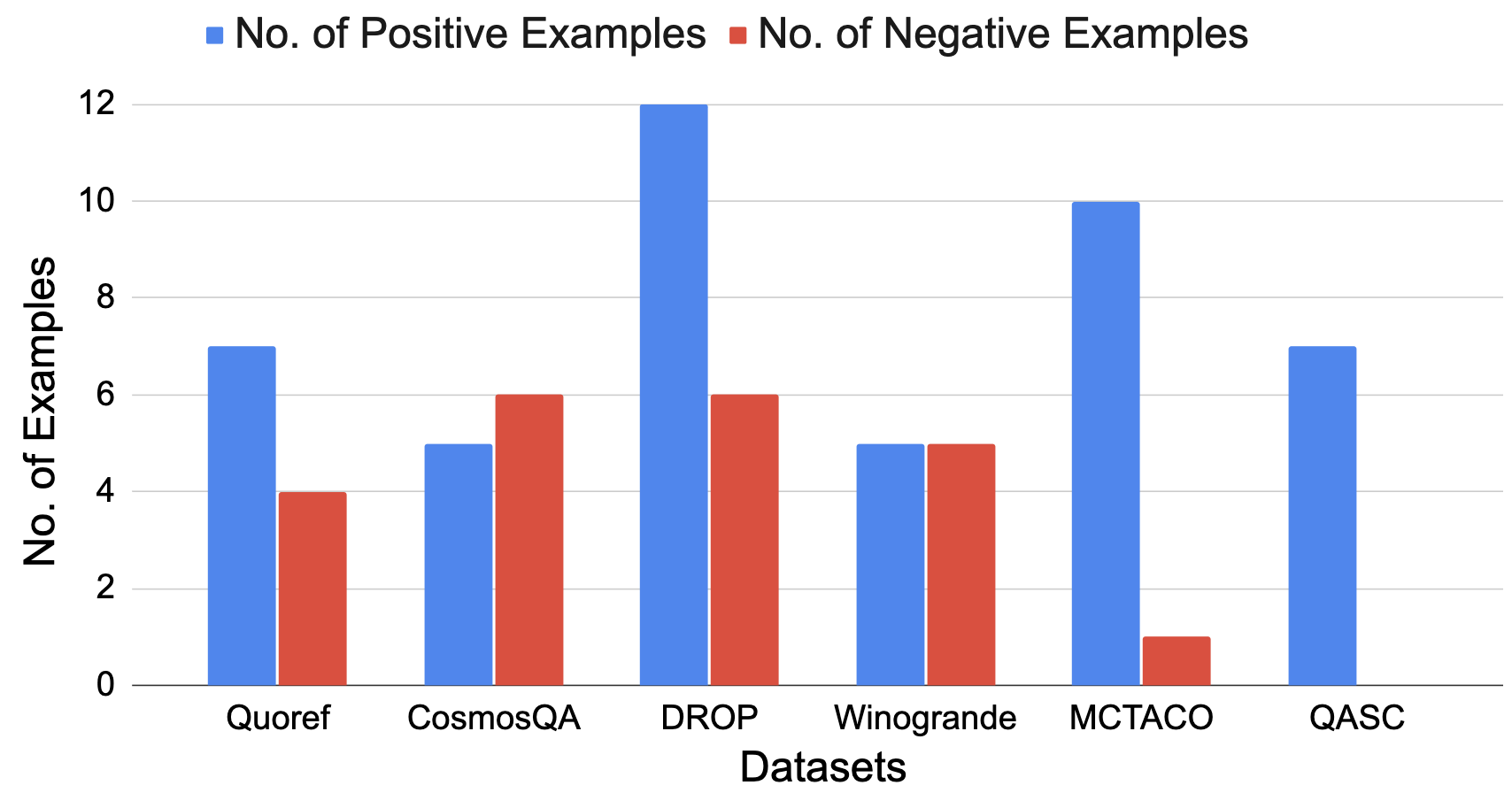}
    \caption{Variation in the number of positive and negative examples}
\label{fig:no. of examples}
\end{figure}

\paragraph{The number of positive/negative examples.} 
Variation in the occurrence of \textsc{Positive} and \textsc{Negative} Examples across datasets has been illustrated in Fig.~\ref{fig:no. of examples}. Only Winogrande provides an equal number of \textsc{Positive} and \textsc{Negative} Examples. 
QASC instructions do not contain any \textsc{Negative} Examples. 
Overall, DROP instructions consist of a relatively higher number of examples than other datasets.

\paragraph{Presence of reasons/suggestions in examples.} All datasets except QASC contain both \textsc{Positive} and \textsc{Negative} Examples. 
However, Quoref is the only dataset to provide \textsc{Reasons} for all the \textsc{Positive} and \textsc{Negative} Examples. There are explanations associated with each of the \textsc{Negative} Examples, but the presence of explanations associated with \textsc{Positive} Examples varies across datasets. Finally, Quoref is the only dataset to provide \textsc{Suggestions} along with the \textsc{Reasons} associated with the \textsc{Negative} Examples.

\subsection{Qualitative Analysis}
\paragraph{Writing Style.} There are significant variation in writing style across the datasets, even among those datasets 
that have the common a objective (e.g., DROP, Quoref and QASC). 
DROP instructions say \textit{"There is an AI running in the background which will also try to answer the question. You won't be able to submit the question if the AI gives the same response."} The writing style in Quoref however is different: \textit{"We also want you to avoid questions that can be answered correctly by someone without actually understanding the paragraph. ..."} 

\paragraph{Information.} We observe that sometimes instructions of a dataset contain information that is relevant to several other datasets, which do not contain similar instruction information. For example, Quoref, DROP and CosmosQA are datasets that are all based on reading comprehension tasks. CosmosQA contains a step in the data creation process asking users to skip passages containing inappropriate or offensive content. This information is also relevant to Quoref and DROP, but is not mentioned in their respective instructions.



\begin{figure}[t]
    \centering
    \includegraphics[scale=0.36,trim=0.1cm 0.1cm 0.1cm 0.1cm]{figures/Task_specification.pdf}
    \caption{Variation in Task Specification: Quoref contains a single line instruction whereas the CosomosQA contains a detailed instruction. QASC on the other hand, contains examples along with instruction.}
    \label{fig:task_specification}
\end{figure}

\paragraph{Hardness.} In a typical crowdsourcing task, certain tasks may be harder than the others, often these are the core tasks, e.g.: question generation, adversarial data creation, etc. Additional information, especially in the form of tips is always helpful in solving these hard tasks. Figure~\ref{fig:task_specification} illustrates that the task of question generation is stated differently in Quoref, CosmosQA, and QASC. QASC mentions an easy and detailed way to create questions, whereas CosmosQA mentions several different attributes of a good quality question. Knowing about the CosmosQA and QASC question generation processes may help with data creation for Quoref and other such question generation tasks, where less additional information is provided regarding question creation.


\subsection{Data Curation Effort}
\label{appendix:subsect:curation}
\begin{figure*}[t]
    \centering
    \includegraphics[scale=0.75,trim=0.7cm 0.5cm 0.5cm 1.5cm]{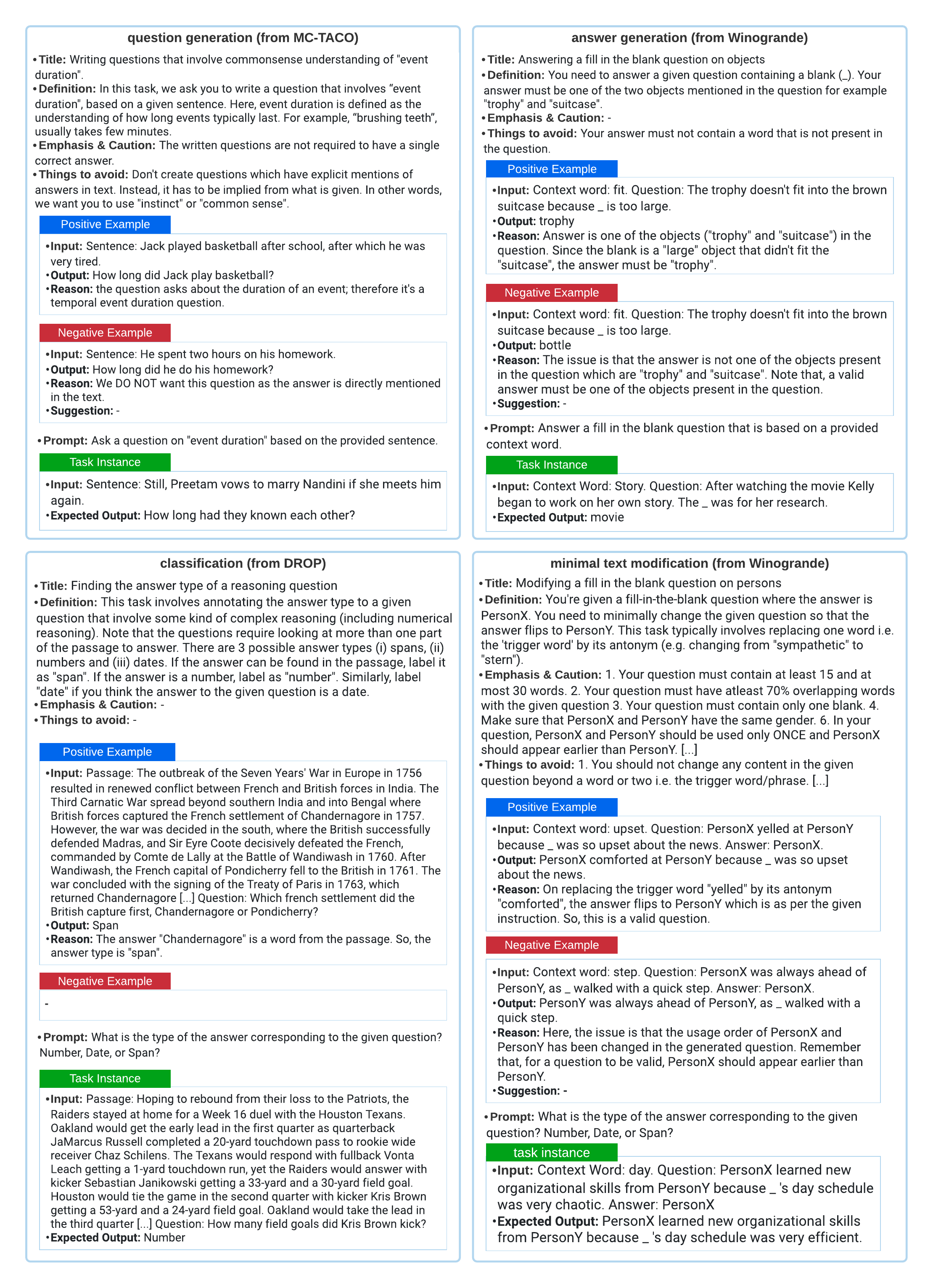}
    \caption{
        Examples from \name{}. 
        Each task follows the schema provided in Fig.~\ref{fig:schema_plate}. 
    }
    \label{fig:examplesfull}
\end{figure*}

Table \ref{tab:datacuration} shows the effort distribution in the data curation process of \name{}. Step-8 which involves parsing instances is the main bottleneck in the data curation process. Table \ref{tab:structure} shows the detailed structure of tasks in \name{}. Fig.~\ref{fig:examplesfull} shows examples of four different tasks in \name{}.

\begin{table}[h]
    \centering
    \footnotesize
    \begin{tabular}{m{0.5cm}p{4.5cm}p{1.5cm}}
        \toprule
        step & task & time per task  \\ 
        \midrule
        1 & Identify crowdsourced dataset and engage with their authors.  & 20-30 mins \\
        2 & Go through the template and understand the task. & 10-15 mins \\ 
        3 & Manually fill fields in the schema with content from the template. & 30-45 mins \\ 
        4 & Iterate over the instructions to ensure their clarity while eliminating the repeated content. Fix writing issue in examples, also typos etc. 
        & 2-3 hrs\\ 
        5 & Create negative examples if not present. Add the missing explanations to the examples. & 1-2 hrs \\ 
        6 & Extract the input/output instances from raw crowdsourcing annotations.  & 0.5-24 hrs \\ 
        7 & Final inspections of the data to verify the data quality 
        & 0.25- 2hrs \\
        \midrule
         & Overall & 6-34 hrs\\
        \bottomrule
    \end{tabular}
    \caption{Steps taken to curate each task in \name{} and their estimated times.
    }
    \label{tab:datacuration}
\end{table}

\begin{table*}
    \centering
    \small
    \begin{adjustbox}{max width=\textwidth}
    \begin{tabular}{llcc}
        \toprule
        task id & title& source dataset & task category\\
        \midrule
1 & task001\_quoref\_question\_generation & Quoref & Question Generation \\
2 & task002\_quoref\_answer\_generation & Quoref & Answer Generation \\
\midrule 
3 & task003\_mctaco\_question\_generation\_event\_duration & MC-TACO & Question Generation \\
4 & task004\_mctaco\_answer\_generation\_event\_duration & MC-TACO & Answer Generation  \\
5 & task005\_mctaco\_wrong\_answer\_generation\_event\_duration & MC-TACO & Incorrect Answer Generation  \\
6 & task006\_mctaco\_question\_generation\_transient\_stationary & MC-TACO & Question Generation  \\
7 & task007\_mctaco\_answer\_generation\_transient\_stationary & MC-TACO & Answer Generation  \\
8 & task008\_mctaco\_wrong\_answer\_generation\_transient\_stationary & MC-TACO & Incorrect Answer Generation  \\
9 & task009\_mctaco\_question\_generation\_event\_ordering & MC-TACO & Question Generation \\
10 & task010\_mctaco\_answer\_generation\_event\_ordering & MC-TACO & Answer Generation  \\
11 & task011\_mctaco\_wrong\_answer\_generation\_event\_ordering & MC-TACO & Incorrect Answer Generation  \\
12 & task012\_mctaco\_question\_generation\_absolute\_timepoint & MC-TACO & Question Generation  \\
13 & task013\_mctaco\_answer\_generation\_absolute\_timepoint & MC-TACO & Answer Generation  \\
14 & task014\_mctaco\_wrong\_answer\_generation\_absolute\_timepoint & MC-TACO & Incorrect Answer Generation  \\
15 & task015\_mctaco\_question\_generation\_frequency & MC-TACO & Question Generation  \\
16 & task016\_mctaco\_answer\_generation\_frequency & MC-TACO & Answer Generation  \\
17 & task017\_mctaco\_wrong\_answer\_generation\_frequency & MC-TACO & Incorrect Answer Generation \\
18 & task018\_mctaco\_temporal\_reasoning\_presence & MC-TACO & Classification  \\
19 & task019\_mctaco\_temporal\_reasoning\_category & MC-TACO & Classification  \\
20 & task020\_mctaco\_span\_based\_question & MC-TACO & Classification  \\
21 & task021\_mctaco\_grammatical\_logical & MC-TACO & Classification  \\
\midrule 
22 & task022\_cosmosqa\_passage\_inappropriate\_binary & Cosmosqa & Classification  \\
23 & task023\_cosmosqa\_question\_generation & Cosmosqa & Question Generation  \\
24 & task024\_cosmosqa\_answer\_generation & Cosmosqa & Answer Generation  \\
25 & task025\_cosmosqa\_incorrect\_answer\_generation & Cosmosqa & Incorrect Answer Generation \\
\midrule 
26 & task026\_drop\_question\_generation & DROP & Question Generation  \\
27 & task027\_drop\_answer\_type\_generation & DROP & Classification  \\
28 & task028\_drop\_answer\_generation & DROP & Answer Generation \\
\midrule 
29 & task029\_winogrande\_full\_object & Winogrande & Minimal Text Modification  \\
30 & task030\_winogrande\_full\_person & Winogrande & Minimal Text Modification  \\
31 & task031\_winogrande\_question\_generation\_object & Winogrande & Question Generation  \\
32 & task032\_winogrande\_question\_generation\_person & Winogrande & Question Generation  \\
33 & task033\_winogrande\_answer\_generation & Winogrande & Answer Generation  \\
34 & task034\_winogrande\_question\_modification\_object & Winogrande & Minimal Text Modification  \\
35 & task035\_winogrande\_question\_modification\_person & Winogrande & Minimal Text Modification \\
\midrule 
36 & task036\_qasc\_topic\_word\_to\_generate\_related\_fact & QASC & Minimal Text Modification \\
37 & task037\_qasc\_generate\_related\_fact & QASC & Minimal Text Modification  \\
38 & task038\_qasc\_combined\_fact & QASC & Minimal Text Modification  \\
39 & task039\_qasc\_find\_overlapping\_words & QASC & Verification  \\
40 & task040\_qasc\_question\_generation & QASC & Question Generation  \\
41 & task041\_qasc\_answer\_generation & QASC & Answer Generation \\
42 & task042\_qasc\_incorrect\_option\_generation & QASC & Incorrect Answer Generation \\
\midrule 
43 & task043\_essential\_terms\_answering\_incomplete\_questions & Essential Terms & Answer Generation  \\
44 & task044\_essential\_terms\_identifying\_essential\_words & Essential Terms & Verification  \\
\midrule 
45 & task045\_miscellaneous\_sentence\_paraphrasing & Miscellaneous & Minimal Text Modification  \\
46 & task046\_miscellaenous\_question\_typing & Miscellaenous & Classification  \\
47 & task047\_miscellaenous\_answering\_science\_questions & Miscellaenous & Answer Generation \\
\midrule 
48 & task048\_multirc\_question\_generation & MultiRC & Question Generation  \\
49 & task049\_multirc\_questions\_needed\_to\_answer & MultiRC & Classification  \\
50 & task050\_multirc\_answerability & MultiRC & Classification  \\
51 & task051\_multirc\_correct\_answer\_single\_sentence & MultiRC & Answer Generation  \\
52 & task052\_multirc\_identify\_bad\_question & MultiRC & Classification  \\
53 & task053\_multirc\_correct\_bad\_question & MultiRC & Minimal Text Modification  \\
54 & task054\_multirc\_write\_correct\_answer & MultiRC & Answer Generation  \\
55 & task055\_multirc\_write\_incorrect\_answer & MultiRC & Incorrect Answer Generation  \\
56 & task056\_multirc\_classify\_correct\_answer & MultiRC & Classification  \\
57 & task057\_multirc\_classify\_incorrect\_answer & MultiRC & Classification  \\
58 & task058\_multirc\_question\_answering & MultiRC & Answer Generation  \\
\midrule 
59 & task059\_ropes\_story\_generation & ROPES & Minimal Text Modification  \\
60 & task060\_ropes\_question\_generation & ROPES & Question Generation  \\
61 & task061\_ropes\_answer\_generation & ROPES & Answer Generation \\
        \bottomrule
    \end{tabular}
    \end{adjustbox}
    \caption{Detailed set of tasks included in \name{}}
    \label{tab:structure}
\end{table*}

\clearpage
\onecolumn

\changed{
\subsection{Qualitative Comparison to PromptSource}
\label{subsec:promptsource}
We provide a comparison between our proposed dataset and PromptSource~\cite{sanh2021multitask}. 
PromptSource tasks are mainly focused on the common NLP downstream tasks (such as question-answering, coreference, NLI, etc).  
However, since we create tasks from various steps (including the intermediate steps) in a data creation process, our instructions contain a broader variety of tasks. For example, tasks for chaining facts (task 38; Table~\ref{tab:structure}), question typing (task 27; Table~\ref{tab:structure}) or detecting inappropriate content (task 22; Table~\ref{tab:structure}) are unique additions in \name{}. 
Additionally, since our instructions were originally written by various researchers targeted for crowdworkers, they are elaborate and contain the complete definition of each task. 
This is somewhat evident from observation that GPT3 leads to higher performance on our instructions (Table~\ref{tab:prompt:source:gpt3:eval}). 
Last but not least, since we represent the instructions in a structured format, we are able to ablate various elements of the instructions (definition, negative/positive examples, etc.) and empirically quantify their contributions (\S\ref{sec:experiments}).  
}

\begin{table}[h]
    \centering
    \small
    \begin{tabular}{clcc}
        \toprule 
        Task & Model & PromptSource & \name{} \\
        \midrule
        \multirow{2}{*}{ Quoref QA (002) } & GPT3-Instruct & 43 & {\bf 47} \\
            & GPT3 & 2 & {\bf 13} \\
        \multirow{2}{*}{ DROP QA (028) } & GPT3-Instruct & 6 & {\bf 10} \\
            & GPT3 & 2 &  {\bf 3} \\
        \bottomrule
    \end{tabular}
    \caption{ 
        Comparing zero-shot performance of GPT3 on our instructions vs. PromptSource. 
        The instructions curated in this work, despite being lengthier, lead to higher performance. 
        }
    \label{tab:prompt:source:gpt3:eval}
\end{table}

\begin{table*}[h]
    \centering
    \includegraphics[scale=0.88,trim=1.4cm 13.4cm 1.2cm 1.85cm,clip=true]{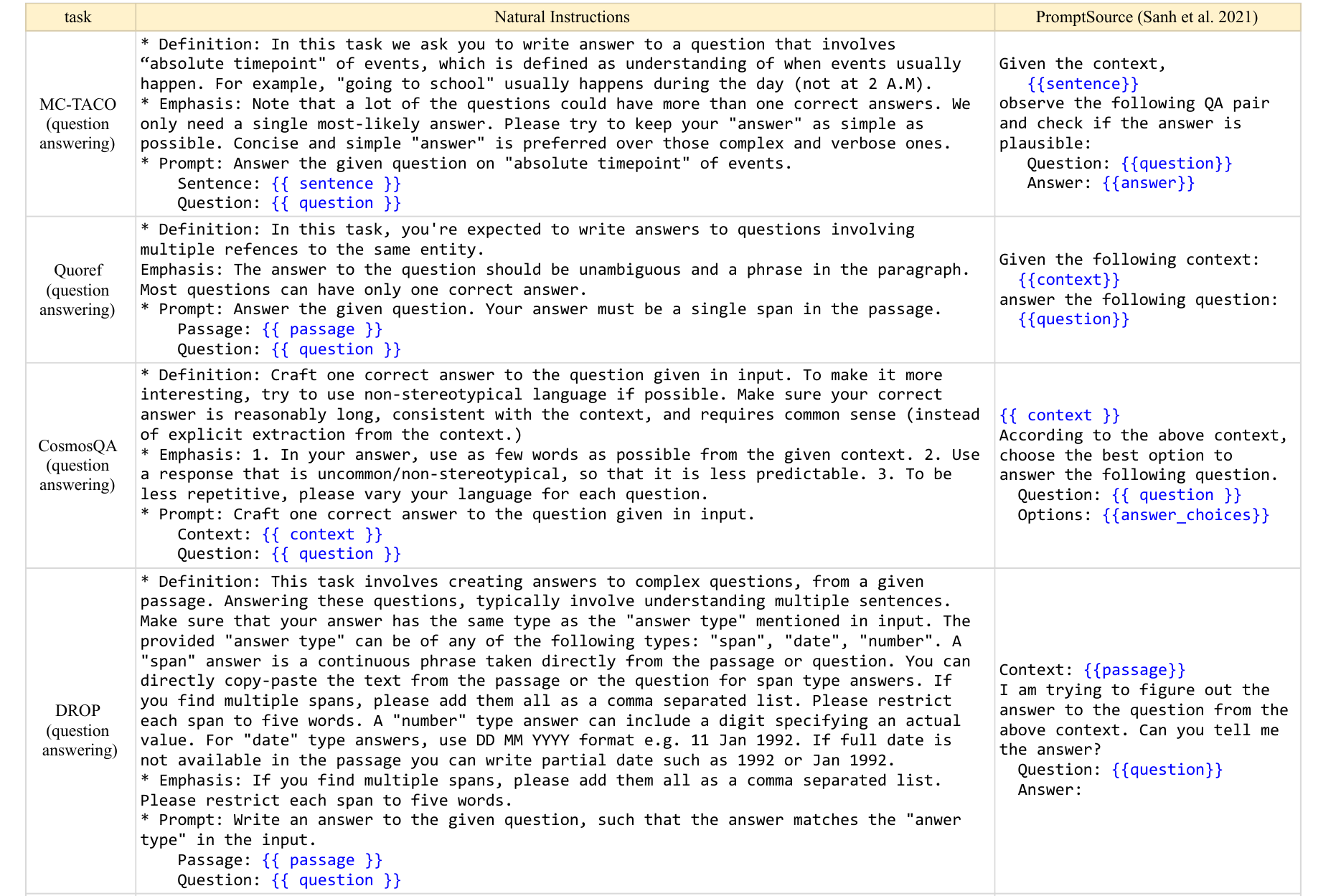}
    \caption{Qualitative comparison of the task instructions for several shared tasks among \name{} and PromptSource~\cite{sanh2021multitask}.}
    \label{tab:prompt:source}
\end{table*}

\twocolumn
\clearpage

\section{Building Baselines for \name{}}
In this section, we provide several details on the baselines included in our work. 

\subsection{Encoding of the instructions}
\label{appendix:subsect:encoding}

According to our schema (\S\ref{subsec:schema}), each instruction $I_t$ for the $t$-th task is a set that contains the following fields:
$$
I_t = \setOf{ 
        \I{t}{title},  
        \I{t}{def.},  
        \I{t}{avoid},  
        \I{t}{emph.},  
        \I{t}{prompt},
        \I{t}{pos. ex.},
        \I{t}{neg. ex.}
    }
$$

To feed the instances to LMs, we first encoder them into plain text. 
Let $enc(I, x)$ define a function that maps a given instruction $I$ and input instance $x$ to plain text. 
Evidently, there are many choices for this function. 
In our study, we consider the following encodings: 

\paragraph{\textsc{No-instructions} encoding.}
This encoding is the conventional paradigm where no instructions exist: 

\begin{equation} \label{eq1}
    \small
    \begin{split}
        enc(I_t, x)  := & \mathtt{\small input:} \; x \\ 
        & \mathtt{\small output:}  \textnormal{''}
    \end{split}
\end{equation}

\paragraph{\textsc{prompt} encoding.}
In this encoding, we append the prompt message before the input:

\begin{equation} \label{eq2}
    \small
    \begin{split}
        enc(I_t, x)  := & \mathtt{\small Prompt:} \; \I{t}{prompt}  \\  
        & \mathtt{\small input:} \; x \\ 
        & \mathtt{\small output:}  \textnormal{''}
    \end{split}
\end{equation}

\paragraph{\textsc{Prompt + Definition} encoding.}
In this encoding, the prompt message and the task definition appear before the input: 
\begin{equation} \label{eq3}
    \small
    \begin{split}
        enc(I_t, x)  := &  \textnormal{``}\mathtt{\small Definition:} \;  \I{t}{def.}  \\ 
        & \mathtt{\small Prompt:} \;  \I{t}{prompt}  \\  
        & \mathtt{\small input:} \; x \\
        & \mathtt{\small output:}  \textnormal{''}
    \end{split}
\end{equation}
Intuitively, this encoding is more informative and more complex than ``prompt'' encoding. 

\paragraph{\textsc{Full Instructions} encoding.}
This encoding contains all the instruction content: 
\begin{equation} 
    \label{eq4}
    \small
    \begin{split}
        enc(I_t, x)  := 
        &  \textnormal{``}\mathtt{\small Definition:} \;  \I{t}{def.}  \\ 
        & \mathtt{\small Prompt:} \;  \I{t}{prompt}  \\  
        & \mathtt{\small Things \; to \; Avoid:} \;  \I{t}{avoid.}  \\  
        & \mathtt{\small Emphasis \& Caution:} \;  \I{t}{emph.}  \\  
        & \textnormal{``}\mathtt{\small Negative Example1-}   \\ 
        & \hspace{0.7cm} \mathtt{\small input:} \; \I{t}{pos. ex.}\mathtt{(input)}  \\  
        & \hspace{0.7cm} \mathtt{\small output:} \; \I{t}{pos. ex.}\mathtt{(output)}  \\  
        & \hspace{0.7cm} \mathtt{\small reason:} \; \I{t}{pos. ex.}\mathtt{(reason)}  \\  
        & \mathtt{\small Negative Example2-}   \\ 
        & \hspace{0.7cm} \mathtt{\small \hdots }  \\ 
        & \textnormal{``}\mathtt{\small Positive Example1-}   \\ 
        & \hspace{0.7cm} \mathtt{\small input:} \; \I{t}{pos. ex.}\mathtt{(input)}  \\  
        & \hspace{0.7cm} \mathtt{\small output:} \; \I{t}{pos. ex.}\mathtt{(output)}  \\  
        & \hspace{0.7cm} \mathtt{\small reason:} \; \I{t}{pos. ex.}\mathtt{(reason)}  \\  
        & \mathtt{\small Positive Example2-}   \\ 
        & \hspace{0.7cm} \mathtt{\small \hdots }  \\ 
        & \mathtt{\small input:} \; x \\ 
        & \mathtt{\small output:}  \textnormal{''}
    \end{split}
\end{equation}

where $enc_{\textnormal{ex}} (I_t)$ is an alternating encoding positive and negative examples. We include as many examples as possible, before exceeding the input limit. 

\paragraph{\textsc{Positive Examples} encoding.}
This encoding contains only positive examples of the subtask (no task description, etc).  

\begin{equation} 
    \label{eq5}
    \small
    \begin{split}
        enc(I_t, x)  := 
        & \hspace{0.7cm} \mathtt{\small input:} \; \I{t}{pos. ex.}\mathtt{(input)}  \\  
        & \hspace{0.7cm} \mathtt{\small output:} \; \I{t}{pos. ex.}\mathtt{(output)}  \\  
        & \hspace{0.7cm} \mathtt{\small \hdots }  \\ 
        & \mathtt{\small input:} \; x \\ 
        & \mathtt{\small output:}  \textnormal{''}
    \end{split}
\end{equation}
Such example-only have been used in several recent studies in the field~\cite{zhao2021calibrate}. 

\clearpage
\onecolumn

\twocolumn

\section{Analysis on Baseline Results}
\label{sec:appendix:banalysis}

\changed{
\subsection{Comparison to Raw Instructions}
\label{subsec:efratlevycomparison}
We seek to understand the value of breaking the tasks into sub-tasks and mapping them into our proposed schema (\S\ref{sec:mapping}). 
We compute performance of raw instructions (first sub-task of four datasets), 
in the same vein as 
\citep{efrat2020turking}'s setup. 
We compare this to our  \textsc{Full Instruction - neg examples} encoding. 
The results in Table~\ref{tab:comparison:raw:instructions} indicate that GPT3 leads to higher performance with our encoding (2nd row) compared to raw instructions (first row). 
Weak performance of LMs on raw instructions aligns with \citep{efrat2020turking}'s finding that ``language model performs poorly''. 

\newcolumntype{R}[2]{%
    >{\adjustbox{angle=#1,lap=\width-(#2)}\bgroup}%
    l%
    <{\egroup}%
}
\newcommand*\rot{\multicolumn{1}{R{30}{1em}}}

\begin{table}[h]
    \small
    \centering
    \begin{tabular}{ccccc}
         \toprule
          & \rot{Quoref} & \rot{MCTaco} & \rot{CosmosQA} & \rot{QASC}  \\
         \midrule
         \makecell{raw instructions} & 12.5 & 5.00 & 6.9 & 3.7  \\
         \makecell{our schema} &       25.8 & 42.6 & 17.7 & 51.3 \\ 
         \bottomrule
    \end{tabular}
    \caption{Comparing GPT3 performance on raw crowdsourcing instructions vs. our encoding. All numbers are ROUGE-L.} 
    \label{tab:comparison:raw:instructions}
\end{table}

This might be partly due to the verbose language of the raw instructions: 
the average length of the raw instructions is $2.5k$ tokens, in comparison to $950$ tokens for our encoding. 
While repetition often helps human understanding, concise instructions seem to be more effective for computers. 
}

\clearpage

\end{document}